\documentclass[a4paper,fleqn]{cas-dc}
\usepackage[ruled,linesnumbered]{algorithm2e}
\usepackage{caption}
\usepackage{amsmath}
\usepackage{hyperref}
\hypersetup{colorlinks=true, citecolor=blue, linkcolor=blue, urlcolor=blue}

\usepackage[numbers]{natbib}
\usepackage{float}
\def\tsc#1{\csdef{#1}{\textsc{\lowercase{#1}}\xspace}}
\tsc{WGM}
\tsc{QE}
\usepackage[switch]{lineno}
\begin{document}
%\linenumbers
\let\WriteBookmarks\relax
\def\floatpagepagefraction{1}
\def\textpagefraction{.001}
\let\printorcid\relax 

% \shorttitle{<short title of the paper for running head>} 
%\shorttitle{MDCGA-Net: Multi-Scale Direction Context-Aware Network with Global Attention for Building Extraction from Remote Sensing Images}    

% \shortauthors{<short author list for running head>}
%\shortauthors{Penghui Niu et al.}

\title[mode = title]{M3S-Net: Multimodal Feature Fusion Network Based on Multi-scale Data for Ultra-short-term PV Power Forecasting}

\author[1]{Penghui Niu}
\ead{qingxinqazxsw@163.com} 
\author[2]{Taotao Cai}
\ead{taotao.cai@unisq.edu.au}
\author[3]{Suqi Zhang}
\ead{suqizhang@tjcu.edu.cn}
\cormark[1]
\author[1,4]{Junhua Gu}
\ead{jhguhebut@163.com} 
\cormark[1]
\author[1,4]{Ping Zhang}
\ead{zhangping@hebut.edu.cn}
\author[5]{Qiqi Liu}
\ead{qiqi6770304@gmail.com}

%\ead[url]{www.cvr.cc,www.tug.org.in}
%\credit{Conceptualization of this study, Methodology, Software}
\author[6]{Jianxin Li}
\ead{jianxin.li@ecu.edu.au}
%\author[2,3]{T. Rishi Nair}[role=Co-ordinator, suffix=Jr]
%\ead{rishi@sayahna.org}
%\ead[URL]{www.sayahna.org}
%\credit{Data curation, Writing - Original draft preparation}

%\author[1,3]{Karl Berry}
%\cormark[2] 
%\fnmark[1,3]
%\ead{karl@freefriends.org} 
%\ead[URL]{www.tug.org}

\address[1]{School of Artificial Intelligence, Hebei University of Technology, Tianjin 300401, China}
\address[2]{University of Southern Queensland, Toowoomba 487-535, Australia}
\address[3]{School of Information Engineering, Tianjin University of Commerce, Tianjin 300134, China}
\address[4]{Hebei Province Key Laboratory of Big Data Calculation, Hebei University of Technology, Tianjin 300401, China}
\address[5]{Trustworthy and General AI Lab, School of Engineering, Westlake University, Hangzhou, 310030, China}
\address[6]{Discipline of Business Systems and Operations, School of Business and Law, Edith Cowan University, Joondalup, WA 6027, Australia}
\cortext[1]{Corresponding author}

\begin{abstract}
The inherent intermittency and high-frequency variability of solar irradiance, particularly during rapid cloud advection, present significant stability challenges to high-penetration photovoltaic (PV) grids. Although multimodal forecasting has emerged as a viable mitigation strategy, existing architectures predominantly rely on shallow feature concatenation and binary cloud segmentation, thereby failing to capture the fine-grained optical features of clouds and the complex spatiotemporal coupling between visual and meteorological modalities. To bridge this gap, this paper proposes M3S-Net, a novel multimodal feature fusion network based on multi-scale data for ultra-short-term PV power forecasting. First, a multi-scale partial channel selection network leverages partial convolutions to explicitly isolate the boundary features of optically thin clouds, effectively transcending the precision limitations of coarse-grained binary masking. Second, a multi-scale sequence to image analysis network employs Fast Fourier Transform (FFT)-based time-frequency representation to disentangle the complex periodicity of meteorological data across varying time horizons. Crucially, the model incorporates a cross-modal Mamba interaction module featuring a novel dynamic “C-matrix swapping” mechanism. By exchanging state-space parameters between visual and temporal streams, this design conditions the state evolution of one modality on the context of the other, enabling deep structural coupling with linear computational complexity, thus overcoming the limitations of shallow concatenation. Experimental validation on the newly constructed fine-grained PV power dataset (FGPD) demonstrates that M3S-Net achieves a mean absolute error (MAE) reduction of 6.2$\%$ and a R-squared ($\mathrm { R^{2}}$) of 0.964 in 10-minute forecasts compared to state-of-the-art (SOTA) baselines. These results confirm that deep cross-modal interaction and micro-physical cloud modeling are instrumental in mitigating ramp-rate risks in real-time grid operations. The dataset and source code will be available at https://github.com/she1110/FGPD.
\end{abstract}


%\item We propose a novel building extraction network structure, MDCGA-Net, which focuses on multi-scale contextual information and global attention feature extraction to improve segmentation results.
%\item We propose the MDCM to assign weights to building boundary detail information by introducing an attention operation with directional information into multi-scale contextual features, and the convolution operation is responsible for adaptive feature aggregation.
%\item We propose the GAGM to extract the discriminant features from all the channels of the features of different scales. The operation of global attention flow combines the attention weights between different feature layers to obtain the correlation between global features.

% Keywords

\maketitle

% Main text
\section{Introduction}\label{sec:intro}
The global transition toward carbon-neutral energy paradigms has precipitated an exponential expansion in solar photovoltaic capacity. While this proliferation is underpinned by the inherent modularity and precipitous decline in the levelized cost of energy of PV technologies, their large-scale integration introduces profound stochasticity into modern power grids \cite{Intro1}. Unlike dispatchable fossil-fuel generation, PV output is intrinsically coupled with atmospheric dynamics, exhibiting volatility across spatiotemporal scales ranging from seasonal cycles to sub-second fluctuations \cite{Intro2}. As PV penetration rates approach critical thresholds within national grids, the stability of the power system becomes increasingly susceptible to this variability \cite{Intro3}. Consequently, the operational imperative has shifted from merely predicting diurnal cycles to mitigating “ramp events,” which are the rapid, high-magnitude power gradients primarily induced by cloud advection \cite{Intro4}. In this context, ultra-short-term forecasting (with horizons spanning minutes to one hour) transcends its traditional role as an ancillary service, emerging as a foundational prerequisite for real-time economic dispatch, frequency regulation, and the efficient operation of ancillary service markets \cite{Intro5}.

Conventional forecasting frameworks rely predominantly on numerical weather prediction (NWP) models or statistical persistence methods. While NWP models provide essential synoptic-scale context for day-ahead planning, their inherent spatiotemporal resolution limitations (typically kilometers and hours) and high latency render them insufficient for capturing the localized, high-frequency transients characteristic of ultra-short-term horizons. Similarly, statistical approaches predicated solely on historical autocorrelation, such as autoregressive integrated moving average or early recurrent neural networks (RNNs), establish a baseline but lack the capacity to anticipate exogenous meteorological disturbances, effectively operating in a purely reactive modality \cite{Intro6, Intro7, Intro8}. These models often fail to capture the abrupt phase shifts caused by rapid cloud movement, leading to significant prediction errors during ramp events.

To overcome these limitations, recent scholarships have pivoted toward deep learning architectures capable of ingesting high-dimensional data. Singlemodal approaches utilizing historical power data, such as bi-directional gated recurrent units (Bi-GRU), have demonstrated that capturing temporal dependencies in both forward and reverse directions can significantly reduce mean squared error (MSE) compared to standard ensembles \cite{Intro9}. Concurrently, the integration of ground-based cloud images (GCI) has established a parallel research trajectory focused on visual irradiance mapping. Early works explicitly modeled cloud kinematics via optical flow, effectively bifurcating the SOTA into two distinct domains: temporal sequence modeling and computer vision-based GCI analysis \cite{Intro10}.

Despite the sophistication of these individual approaches, a critical research gap persists: the inability of existing models to effectively synthesize the complex, multi-scale dynamics of PV generation when influenced jointly by meteorological trends and chaotic cloud dynamics. First, while GCI offers a theoretical solution to the latency inherent in time-series data, current implementations are severely constrained by coarse-grained feature extraction. By reducing complex cloud structures to binary sky-cloud masks, existing visual models fail to account for fine-grained attributes such as cloud scale, color, and radiation source interactions. This oversight introduces an irreducible error floor in irradiance prediction, particularly during the passage of translucent or multi-layered cloud formations. Second, to address single-modal limitations, the field has moved toward multimodal fusion. However, contemporary fusion architectures typically extract spatial and temporal features in isolation, employing “late fusion” strategies that concatenate representations only at the final decision layer. This isolation precludes deep structural coupling, preventing the internal state of one modality from dynamically conditioning the processing of the other. Furthermore, few models explicitly address the hierarchical periodicity inherent in the data, where rigid diurnal cycles are overlaid with stochastic high-frequency noise. There is a distinct lack of architectures capable of simultaneously handling fine-grained visual segmentation, multi-scale temporal periodicity, and deep, low-latency cross-modal fusion.

To resolve these distinct challenges, this research proposes M3S-Net (MultiModal feature fusion network based on Multi-Scale data), a novel architecture designed to achieve high-fidelity ultra-short-term forecasting by unifying fine-grained visual perception with robust multi-scale temporal modeling. Specifically, we introduce three branches of methodological innovations to resolve the identified gaps. Fine-grained visual extraction branch: We propose the Multi-scale Network integrating Partial Channel Selection (MPCS-Net). Unlike binary segmentation methods, MPCS-Net incorporates a Spatial-Channel Selection Mechanism (SCSM) and partial convolutions. This architecture is explicitly designed to model the translucent boundaries and internal texture of clouds, effectively capturing the varying optical depths that drive ramp events. Multi-scale temporal imaging branch: To handle the complex periodicity of power data, we introduce the multi-scale Sequence to Image analysis Network based on FFT and Retractable attention (SIFR-Net). SIFR-Net utilizes FFT to convert 1D time-series into 2D temporal images, enabling the application of a Retractable Attention (RA) mechanism. This allows the model to dynamically zoom its receptive field, capturing both high-frequency local fluctuations and low-frequency seasonal trends via spectral-temporal disentanglement. Cross-modal mamba fusion branch: We depart from standard Transformer or concatenation fusion by implementing a Multimodal Interaction Fusion (MMIF) module based on Mamba. This module employs a novel “C-matrix swapping” mechanism, where the state-space parameters of the visual stream are exchanged with the temporal stream. This forces the model to decode the visual information conditioned on the temporal state, and vice versa, achieving a true synthesis of modalities with linear computational complexity. Experimental validation on the newly released FGPD and public solar radiation research laboratory (SRRL) datasets demonstrates that M3S-Net significantly outperforms SOTA baselines. These results confirm that deep cross-modal interaction and fine-grained cloud modeling are decisive factors in mitigating the risks of ultra-short-term PV volatility. The main contributions of this paper are summarized as follows.
\begin{enumerate}
	\item{A novel cross-modal mamba fusion framework, M3S-Net, is proposed to overcome the limitations of shallow concatenation in multimodal forecasting with a hierarchical architecture. By introducing a cross-modal Mamba interaction mechanism with a novel "C-matrix swapping" strategy, we enable deep cognitive coupling between visual and temporal modalities. This allows the model to dynamically condition its visual interpretation on the temporal context with linear computational complexity.}
	\item{A multi-scale partial channel selection network, MPCS-Net, is designed to address the inadequacy of binary cloud masks. By integrating a SCSM with partial convolutions, this network explicitly models the translucent boundaries and optical features of clouds. This innovation enables the precise extraction of physically relevant texture features that are critical for quantifying solar irradiance attenuation.}
	\item{A multi-scale sequence to image analysis network, SIFR-Net, based on FFT and retractable attention is introduced to disentangle the complex periodicity of sequence data. By utilizing a sequence-to-image transformation, this module dynamically adjusts its receptive field to capture both high-frequency fluctuations (local cloud transients) and low-frequency trends (diurnal cycles), effectively mitigating performance degradation in longer-horizon forecasts.}
	\item{The introduction and public release of the FGPD dataset. As a significant contribution to the community, FGPD introduces meteorological and power plant data based on the CSRC dataset. Covering various classic weather conditions such as sunny, cloudy, and overcast, this dataset ensures diversity and establishes a more practical and challenging foundation for advancing ultra-short-term PV power forecasting.}
\end{enumerate}

The rest of this paper is organized as follows. Section~\ref{sec:related} reviews related studies pertaining to singlemodal and multimodal PV power forecasting. Section~\ref{sec:method} details the structural design of the proposed M3S-Net. Section~\ref{sec:setting} describes the experimental settings and dataset characteristics. Section~\ref{sec:exp} presents the performance evaluation and comparative analysis. Finally, Section~\ref{sec:conclusion} concludes the paper.

\section{Related Works} \label{sec:related}
PV power forecasting is a prerequisite for ensuring grid stability, optimizing economic dispatch, and maximizing the hosting capacity of renewable energy systems. As global PV penetration accelerates, the precision of ultra-short-term forecasting has emerged as a critical research imperative. Traditional methodologies, predominantly reliant on singlemodal data sources such as historical power sequences or NWP outputs, often fail to capture the high-frequency variability in irradiance induced by rapid cloud dynamics. Consequently, these conventional models struggle to anticipate “ramp events,” thereby threatening grid inertia and operational security.

In recent years, facilitated by advancements in deep learning technologies and multimodal data fusion methodologies, researchers have begun to synthesize multi-source information integrating GCI, meteorological time-series data, and historical power records to augment forecasting precision. This section provides an exhaustive review of existing research progress, categorized into single-modal data forecasting methods, multimodal data fusion forecasting methods, and relevant datasets.

\subsection{Singlemodal Data PV Power Forecasting Methods}
Foundational investigations into PV power forecasting were predicated on unimodal data inputs, utilizing either historical generation records or meteorological parameters. These approaches typically exploit the temporal autocorrelation of power output or the physical correlation between atmospheric variables and irradiance. A significant evolution in capturing temporal dependencies involves the application of Bi-GRU. Dai et al. introduced a hybrid ensemble framework optimized via Bi-GRU, achieving a reduction in MSE of approximately 60.4$\%$ \cite{Intro11}. By processing temporal sequences in both forward and backward directions, this architecture effectively captures bidirectional dependencies often overlooked by standard RNNs. Parallel to architectural innovations, Wang et al. proposed a system integrating feature selection with multi-objective optimization, enhancing model stability \cite{Intro12}. Addressing the limitations of deterministic forecasting in modeling the stochastic “texture” of PV generation, Huang et al. deployed a conditional generative adversarial network. Furthermore, representing a shift toward continuous-time modeling, Huang et al. also introduced a multi-stage attention neural network based on neural ordinary differential equations \cite{Intro13}. This architecture optimizes LSTM networks and TCN to perform fine-grained feature extraction, effectively modeling the continuous evolution of solar irradiance beyond discrete time steps.

Distinct from time-series methods, visual approaches utilize GCI to directly observe the primary driver of PV volatility: cloud cover and motion. By analyzing the sky, these models attempt to bypass the lag inherent in time-series data. Foundational work in this domain focuses on mapping visual features directly to irradiance levels. Trigo-Gonzalez et al. developed artificial neural network and support vector machine models utilizing GCI to predict global horizontal irradiance (GHI) \cite{Intro14}. Moving towards end-to-end deep learning, Feng et al. developed two deep CNN models specifically for intra-hour forecasting using only GCI \cite{Intro15}. To explicitly model the kinematics of clouds, Eglik et al. combined computer vision techniques with deep learning, integrating Shi-Tomasi feature detection, the Lucas-Kanade optical flow method, and LSTM networks for solar radiation prediction \cite{Intro16}. Bridging signal processing and computer vision, Liu et al. proposed a sky image-driven deep decomposition method. This novel framework processes GCI to guide the decomposition of the PV power sequence into trend and seasonal components \cite{Intro17}. Representing the SOTA in spatiotemporal feature fusion, Zang et al. introduced a dual-stream network equipped with a PV-guided attention mechanism \cite{Intro18}.

Despite the significant accuracy achieved by single-modal methods, they often fail to capture the complex, multi-scale dynamics of PV generation, which are influenced jointly by meteorological trends and cloud dynamics. This limitation has necessitated the development of proposed multimodal data fusion techniques. By leveraging both the temporal precision of historical data and the spatial context of sky imagery, these fused models aim to utilize complementary information to enhance forecasting robustness and accuracy.

\subsection{Multimodal Data PV Power Forecasting Methods}
To surmount the inherent limitations of unimodal forecasting methodologies, contemporary research has increasingly pivoted towards multimodal data fusion technologies that synthesize ground-based cloud imagery with meteorological parameters and historical power generation data. The seminal work by Kong et al. established a cornerstone in this domain, proposing hybrid architectures such as CNN-LSTM and ConvLSTM to model cloud movement as a spatiotemporal phenomenon \cite{Kong}. Building upon the unidirectional LSTM approach, Dolatabadi et al. developed a hybrid model combining CNN with Bi-directional LSTM \cite{Dolatabadi}. Moving beyond point forecasting, Terrén et al. introduced a deterministic and Bayesian multi-task deep learning model \cite{Terr}. Caldas et al. pioneered a method that tightly couples ground-based cloud maps with real-time irradiance data \cite{Caldas}. Liu et al. approached the problem from a computer vision perspective, employing Sparse Spatiotemporal Descriptors equipped with spatial pyramid pooling \cite{Liu}.

The most recent wave of literature marks a transition from standard CNN-RNN architectures to advanced formulations involving Transformers, TimesNet, and sophisticated clustering algorithms. Wu et al. proposed a structured approach utilizing a CNN coupled with an image RGB matrix analysis to manage sky diversity \cite{Wu}. They introduced a clustering boundary correction algorithm to classify full-sky images into four distinct meteorological categories. Li et al. addressed the modality gap between 1D time-series and 2D images using TimesNet. The deep features from the ground-based cloud maps (via T2T-ViT) are fused with the temporal features to construct a multimodal feature learning model for ultra-short-term solar irradiance prediction \cite{T2T-ViT}. Dou et al. proposed a multimodal deep clustering architecture that performs collaborative multimodal feature learning and clustering assignment simultaneously \cite{Dou}. A multimodal fusion module integrates the learned features, and a specialized predictor outputs radiance values conditioned on the identified weather cluster Zhu et al. proposed a hybrid model for direct normal irradiance (DNI) prediction \cite{Zhu}. The CNN module incorporates dilated convolution to expand the receptive field, and coordinate attention to focus the spatial locations. Innovative transformation techniques have also been explored to enhance feature extraction. Wei et al. introduced the NUTF method, proposing a technique to convert time-series data into an image mode \cite{Wei}. The fused features from this “time-image” and the actual ground-based cloud image generate GHI using fuzzy logic method. Furthermore, Nie et al. utilized gray-level co-occurrence matrices and local binary patterns to capture texture features \cite{Nie}. These features are combined with classification features from an improved ConvNeXt network and cloud occlusion factors, and then processed by an SVM-Informer model. Ma et al. adopted a stacking ensemble strategy that ensemble integrates random forest and XGBoost with LSTM and GRU \cite{Ma}. Wei et al. proposed a hybrid framework fusing discrete wavelet transform that decomposes the non-stationary power series into multiple frequency components, CNN, and LSTM, optimized by the bald eagle search algorithm \cite{Wei2}. Xiang et al. focused on the changing relationships between weather and power \cite{Xiang}. They proposed a dynamic meteorological correlation integrated hybrid method combining XGBoost and temporal convolutional networks. Shi et al. developed CloudMViT, a hybrid architecture integrating CNNs with MobileViT \cite{Shi}. The specialized cmv2 module efficiently extracts local features while maintaining a low computational footprint, facilitating real-time inference on resource-constrained hardware. Yong et al. introduced the LSEConvNeXt module combining low-rank atrous spatial pyramid pooling with ConvNeXt to significantly improve the pixel-wise segmentation of cloud images \cite{ConvODE-Mixer}. The proposed ConvODE-Mixer framework integrates neural ordinary differential equations with ground-based cloud maps to improve the accuracy of PV power forecasting. Shi et al. proposed a dual-branch architecture. One branch utilizes a CNN to extract local features, while a parallel branch uses a Transformer to extract global features \cite{Shi2}. A dedicated fusion module integrates these outputs to achieve a comprehensive feature extraction.

Despite the effective progress achieved by the aforementioned methods in PV power forecasting, many existing multimodal fusion methodologies rely on simple feature concatenation or linear fusion mechanisms. They often lack deep cross-modal interaction and adaptive multi-scale feature extraction capabilities. Furthermore, very few models explicitly address the hierarchical and periodic characteristics inherent in meteorological and power data.

\subsection{Datasets for Multimodal PV Forecasting}
The establishment of standardized benchmarking datasets has been fundamental to the comparative evaluation of solar forecasting algorithms. These repositories, characterized by their temporal continuity and instrumental diversity, serve as the ground truth for validating model generalization.

The solar radiation research laboratory, operated by the U.S. national renewable energy laboratory (NREL) in Golden, Colorado, provides a foundational benchmark for multimodal analysis \cite{SRRL}. Central to the SRRL’s imaging capabilities is the ASI-16 model. The instrument generates RGB channel sky images with a resolution of 1526 $\times$ 1526 pixels. The standard archival resolution for the dataset is typically set to a 10-minute interval. Complementing the optical data is a suite of meteorological measurements sampled at a significantly higher temporal resolution of 60 seconds. These features include: zenith angle, relative humidity, total cloud cover, ambient temperature, average wind speed, wind direction, precipitation, and snow depth. The Folsom dataset originated from Folsom, California. The dataset captures sky images at a 1-minute sampling interval during daylight hours, utilizing a fisheye camera system that produces images with a resolution of 1536 $\times$ 1536 pixels \cite{Folsom}. The dataset provides 1-minute average values of GHI, derived from two Licor LI-200SZ pyranometers. In addition to irradiance and imagery, the Folsom repository includes a comprehensive array of correlated meteorological variables such as ambient temperature, wind speed, and humidity. The SIRTA dataset is collected at the Site Instrumental de Recherche par Télédétection Atmosphérique observatory in Palaiseau, France \cite{SIRTA}. Data acquisition at SIRTA is performed at a 1-second sampling interval, which is then aggregated to provide 1-minute average values. Sky images at SIRTA are captured at a resolution of 768 $\times$ 1024 pixels with a 1-minute sampling interval.

While public benchmarking datasets facilitate algorithmic development, dedicated engineering datasets are constructed to address specific operational challenges. The Gansu power grid dataset (GPGD) \cite{ConvODE-Mixer} represents a significant contribution to the study of large-scale renewable energy integration in China. Located in a region with high solar resource availability but complex grid dynamics, the GPGD encompasses meteorological features and power generation data from seven distinct PV power stations. Utilized by Ma et al. \cite{Ma}, the dataset is situated in Baoying County, Jiangsu Province, representing a plain/coastal climatic zone distinct from the mountain or high-altitude/desert environments of the other datasets. The dataset integrates three months of continuous meteorological data, ground-based cloud imagery, and PV power generation records. The flat terrain of Baoying contrasts with the mountain dataset, shifting the forecasting focus from topographic effects to pure cloud advection and formation processes driven by frontal systems or local convection.

To support the visual reasoning capabilities of forecasting models, research institutions have developed datasets specifically annotated for segmentation and classification tasks. The whole sky image segmentation (WSISEG) and Singapore whole sky nychthemeron image segmentation (SWINySEG) databases are foundational for pixel-level cloud detection. Comprising 400 annotated whole-sky images, the WSISEG dataset provides pixel-level labels for cloud, sky, and undefined regions \cite{WSISEG}. SWINySEG extends the scope to 6,768 images, incorporating both daytime and nighttime scenes \cite{SWINySEG}. Published by Tianjin Normal University, the ground-based cloud dataset (GCD) and ground-based remote sensing cloud database (GRSCD) focus on cloud type classification rather than just segmentation \cite{Shi}. These datasets provide annotated images covering seven distinct cloud types. Recent advancements have pushed beyond simple classification toward fine-grained attribute analysis, addressing the limitation that binary masks cannot capture the optical depth or physical scale of cloud formations. Constructed by Shi et al. \cite{Shi2}, this dataset addresses the unique challenges of mountain PV stations. A pivotal innovation of this dataset is the reclassification of clouds into five distinct categories based on their radiative characteristics, cloud morphology, and associated meteorological features. The dataset features synchronized acquisition of total irradiance, direct irradiance, wind speed, temperature, humidity, atmospheric pressure, and actual power output.  Introduced by Niu et al. \cite{MPCM-Net}, the CSRC dataset represents a significant leap in granularity. It includes pixel-level annotations for fine-grained cloud attributes, such as scale (size of cloud features), color (spectral properties related to optical depth and water content), and radiation source information.
\begin{figure*}[!t]
	\centering{\includegraphics[width=6.5in]{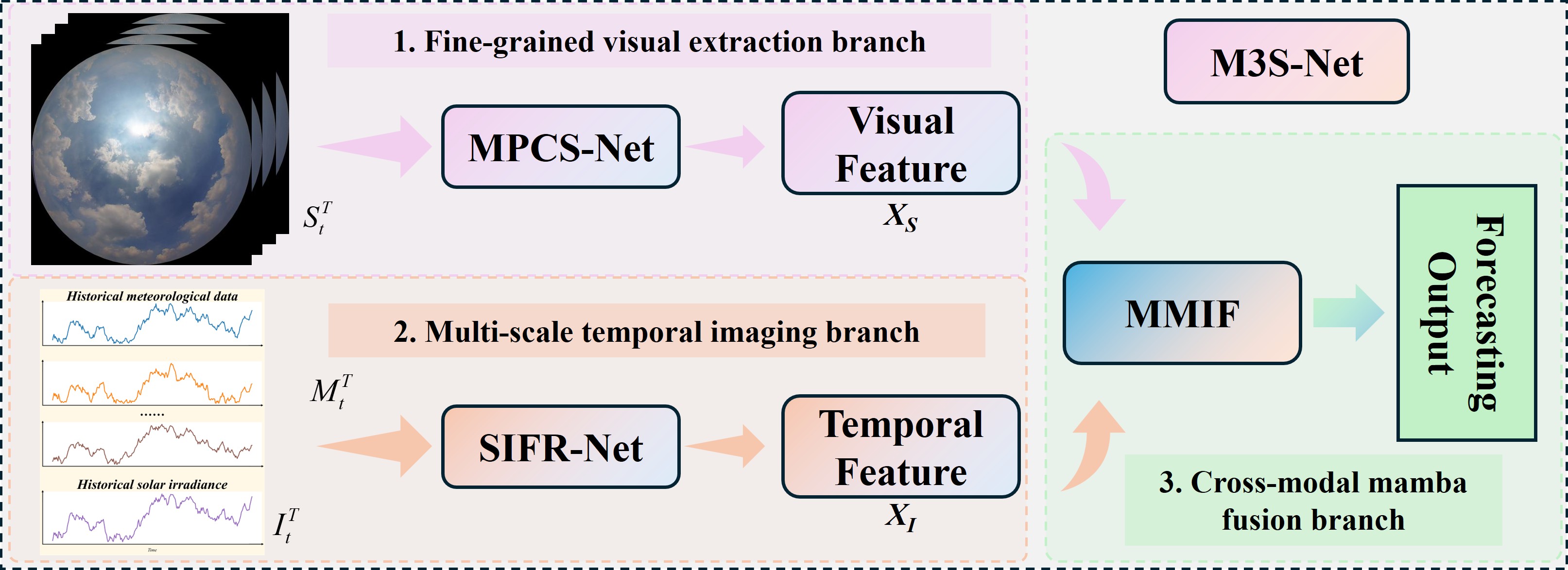}}
	\caption{The overall diagram of the proposed M3S-Net. The framework accomplishes the prediction task through three core components: 1. Fine-grained visual extraction branch; 2. Multi-scale temporal imaging branch; 3. Cross-modal mamba fusion branch.}
	\label{fig_1}
\end{figure*}
Existing ground-based cloud imagery datasets are predominantly limited to a binary classification distinguishing cloud entities from the sky background, thereby neglecting critical features regarding cloud scale, morphology, and chromatic characteristics. Such coarse-grained representations are insufficient for capturing the nuanced impact of distinct cloud types on solar irradiance and fail to meet the stringent requirements of ultra-short-term forecasting models for high-spatiotemporal-resolution cloud information.

\section{Proposed Method}\label{sec:method}
This section delineates the mathematical formulation of the ultra-short-term PV power forecasting problem and presents the comprehensive architecture of the proposed M3S-Net. Subsequent subsections detail the network's three synergistic components: the fine-grained visual extraction branch, the multi-scale temporal imaging branch, and the cross-modal Mamba fusion branch.

%\begin{figure}[t!] 
%\vspace*{-10pt}

\vspace*{-10pt}
%\end{figure}

\subsection{Problem formulation and overall architecture}
Addressed as a sequence-to-sequence regression problem, ultra-short-term PV power forecasting requires the precise mapping of historical multimodal inputs to future power generation over a specified horizon. Achieving this objective necessitates the effective integration of heterogeneous data sources, specifically high-dimensional ground-based cloud imagery and scalar time-series variables. To reconcile the dimensionality disparities between these modalities, the proposed framework incorporates a robust alignment mechanism prior to multi-scale feature extraction. Formally, we define a historical observation window spanning from time step $t$ to $T$. The forecasting task aims to predict the PV power output for the subsequent $\tau$ steps. This process is mathematically formulated as:

\begin{linenomath*}
	\begin{flalign}
		\begin{split} 
			S_{t}^{T}&=\left[S_{t}, S_{t+1}, \ldots, S_{T}\right]\\
			I_{t}^{T}&=\left[I_{t}, I_{t+1}, \ldots, I_{T}\right]\\
			M_{t}^{T}&=\left[M_{t}, M_{t+1}, \ldots, M_{T}\right]
		\end{split} && 
		\label{equ_1}
	\end{flalign}
\end{linenomath*}

\begin{linenomath*}
	\begin{flalign}
		\begin{split}
			I_{T+1}^{T+\tau}&=\left[ I_{T+1}, I_{T+2}, \ldots, I_{T+\tau} \right] \\
			&=F\left( S_{t}^{T}, I_{t}^{T}, M_{t}^{T}, \left\{\theta   \right\}\right)
		\end{split} && 
		\label{equ_2}
	\end{flalign}
\end{linenomath*}

\par\noindent where $ S_{t}^{T}\in\mathbb{R }^{\left( T-t+1\right )\times H\times W\times C}$ represents the ground-based cloud image sequence, with $H$, $W$, and $C$ denoting height, width, and channel depth, respectively. $ I_{t}^{T}\in\mathbb{R }^{T-t+1}$ denotes the historical solar irradiance data. $ I_{T+1}^{T+\tau }\in\mathbb{R }^{\tau}$ denotes the predicted irradiance. $ M_{t}^{T}\in\mathbb{R }^{T-t+1}$ denotes the historical meteorological data encompassing $m$ parameters. The proposed model and its hyperparameters are designated as $F\left( \cdot  \right)$ and $\left \{ \theta  \right \}$, respectively. Fig.~\ref{fig_1} illustrates the overall flowchart of the proposed M3S-Net. The framework accomplishes the prediction task through three core components.

\begin{enumerate}
	\item{Fine-grained visual extraction branch: The MPCS-Net is proposed to extract fine-grained spatial features. It employs a dedicated encoder-decoder architecture to explicitly model multi-scale cloud characteristics and translucent boundaries via a spatial-channel selection mechanism.}
	\item{Multi-scale temporal imaging branch: The SIFR-Net is introduced to handle the inherent periodicity of power generation, this branch converts 1D time-series data into 2D representations using FFT. A retractable attention mechanism is then applied to analyzing multi-scale seasonal and trend features.}
	\item{Cross-modal mamba fusion branch: The MMIF module facilitates the deep interaction of multimodal data. Utilizing a cross-modal mamba architecture, it integrates complementary long-range dependencies by exchanging output projection matrices between modalities, subsequently generating forecasts via a decoding layer.}
\end{enumerate}
\begin{figure*}[!t]
	\centering{\includegraphics[width=6in]{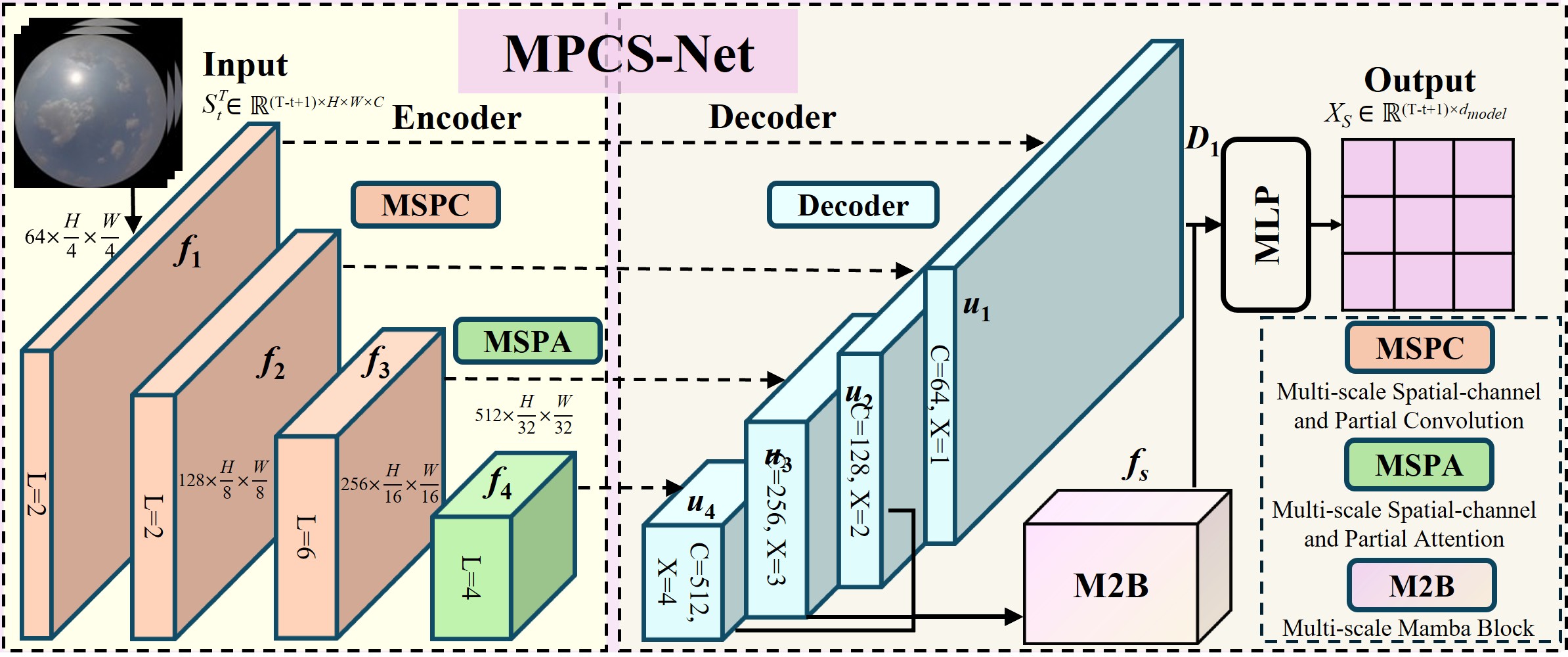}}
	\caption{The structure of the proposed MPCS-Net, which is an Encoder-Decoder architecture. The spatial-channel selection mechanism is embedded within the encoder to form the MSPC and MSPA.}
	\label{fig_2}
\end{figure*}

\subsection{Fine-Grained Visual Extraction Branch Based on MPCS-Net}
Accurate segmentation of GCI is a prerequisite for precise PV power forecasting, as the optical properties and distribution of clouds directly govern solar irradiance variability. Cloud formations exhibit complex multi-scale characteristics and translucent boundaries that challenge conventional segmentation approaches. To address these complexities, we propose the MPCS-Net. Building upon the architecture of MPCM-Net \cite{MPCM-Net}, which explicitly integrated partial attention convolution with Mamba for efficient global context modeling, MPCS-Net introduces a spatial-channel selection mechanism to further refine multi-scale feature extraction.

While the previously established MPCM-Net demonstrated efficacy in capturing boundary details, its reliance on atrous convolution for multi-scale fusion limited the interaction between features at different resolutions. The proposed SCSM overcomes this deficiency by adaptively filtering and enhancing informative features across both spatial and channel dimensions. This mechanism allows the network to dynamically prioritize relevant cloud attributes, such as radiation source intensity and spectral variations, thereby significantly improving segmentation accuracy under complex weather conditions.

\subsubsection{The Encoder of the MPCS-Net}
As depicted in Fig.~\ref{fig_2}, the MPCS-Net encoder comprises four consecutive convolutional stages. Diverging from previous methodologies (MPCM-Net), the SCSM is embedded within all four stages: the first three utilize the multi-scale spatial-channel and partial convolution (MSPC) block, while the final stage employs the multi-scale spatial-channel and partial attention (MSPA) block. This configuration augments the encoder's capacity to adaptively extract multi-scale features and capture inter-channel dependencies. We denote the feature map of the $i$-th layer as $f_{i}$ ($i \in \{1, \dots, 5\}$), with dimensions $ 2^{i-1}\cdot  64\times \left ( H/ 2^{i} \right ) \times \left ( W/ 2^{i} \right ) (i \in[1,2,3,4,5]) $.
\begin{figure*}[!t]
	\centering{\includegraphics[width=6in]{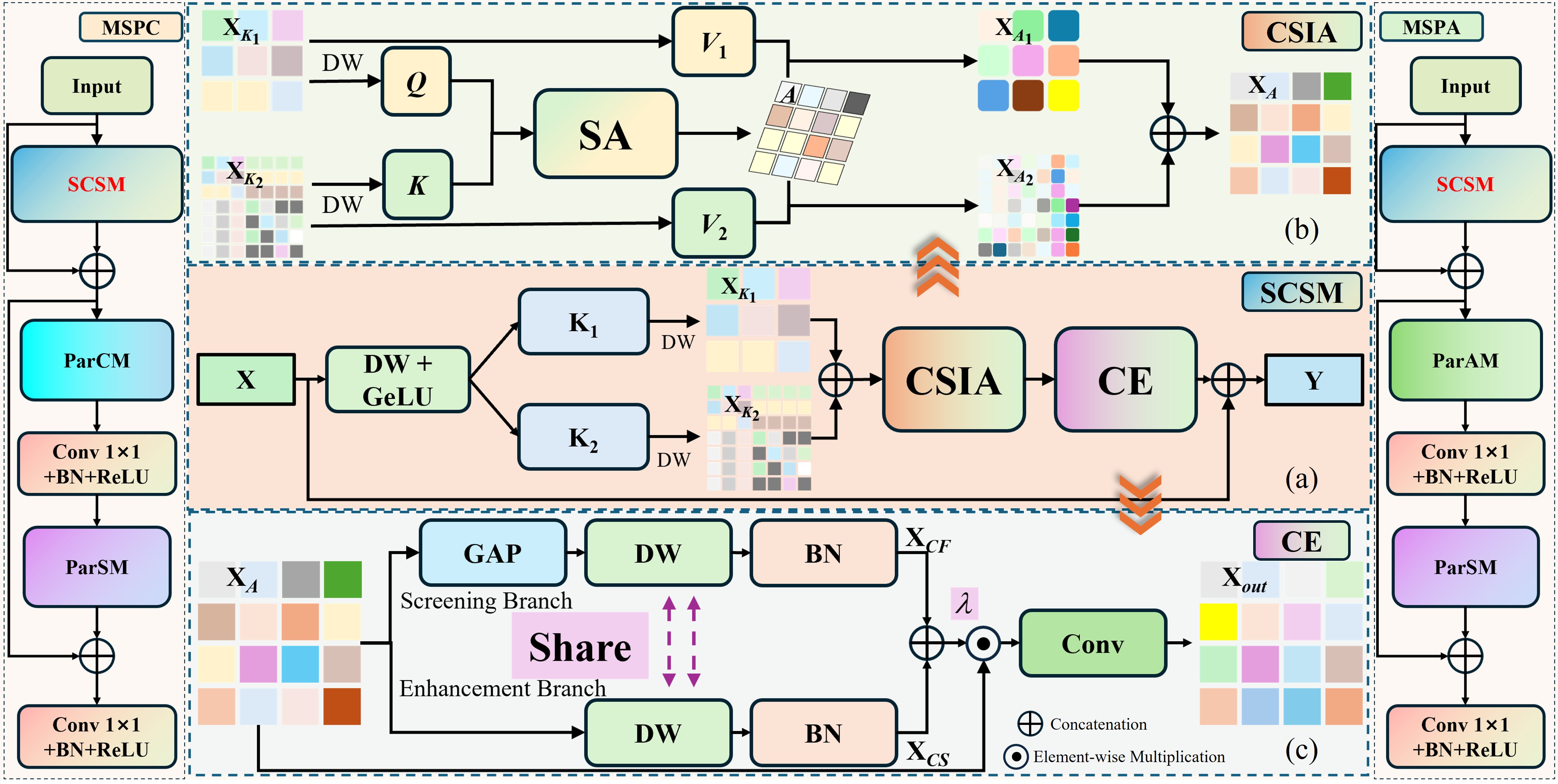}}
	\caption{The structure of the proposed MPCS and MSPA, which is concluding the (a) SCSM. The core modules of the SCSM are (b) CSIA and (c) CE.}
	\label{fig_3}
\end{figure*}
\subsubsection{Spatial-Channel Selective Module}
Accurate segmentation under complex weather conditions relies on the effective capture of multi-scale contextual information. Existing multi-scale atrous convolution pyramid methods, as noted in a previous study, typically concatenate feature maps from different scales, often neglecting the interoperability of partial spatial and channel characteristics. To overcome this limitation, we introduce a SCSM within the MPCM-Net framework. As illustrated in Fig.~\ref{fig_3}, the SCSM introduces a cross-scale interaction attention (CSIA) mechanism. By activating and selecting salient cross-scale feature tokens, this mechanism enhances inter-scale interaction within the encoder. Furthermore, a channel excitation (CE) module \cite{SENet} is integrated to suppress irrelevant channels and capture region-specific patterns via a self-sharing mechanism, thereby efficiently emphasizing critical inter-regional features.

The initial phase of the MPCS-Net pipeline is dedicated to constructing a rich feature representation that balances computational efficiency with semantic depth. As illustrated in Fig.~\ref{fig_3} (a), the input feature tensor $X$ is processed through a depthwise (DW) Convolution layer coupled with a gaussian error linear unit (GeLU) activation function. The strategic selection of DW convolution decouples spatial filtering from channel mixing, substantially reducing floating-point operations (FLOPs) relative to standard convolution. Concurrently, the adoption of GeLU over the traditional ReLU is pivotal. Its smooth, probabilistic approximation of the rectifier facilitates superior gradient flow and enhances non-linear representation capabilities in deep architectures. This configuration ensures high feature extraction efficiency while preserving the non-linear capacity required to model the complex scattering functions of cloud particles. To enable adaptive focus on relevant spatial contexts, essential for extracting cloud clusters of varying scales, the architecture employs explicitly decomposed convolutions. We utilize convolution kernels of differing scales to selectively extract spatial features, generating scale-specific feature maps, $X_{K_{i}}$. This operation is formally expressed as: $X_{K_{i}}=\mathrm {DW}\left ( K_{i} \right )$, where $i$ represents the index of large-scale convolution kernels. Consistent with established methodologies in recent literature \cite{USF-Net}, this strategy exploits the interplay between kernel size and dilation rate. By progressively increasing these parameters, the explicitly decomposed operations synthesize receptive fields of diverse magnitudes, enhancing the network's multi-scale representation capability without the parameter proliferation associated with standard large-kernel convolutions.

MPCS-Net introduces CSIA that is designed to facilitate sophisticated information exchange between feature maps characterized by different receptive fields. As shown in Fig.~\ref{fig_3} (b), we employ an asymmetric projection strategy where the Query ($Q$) and Value ($V_1$) vectors are derived from the low-scale feature map $X_{K_{1}}$, while Key ($K$) and Value ($V_2$) vectors are derived from the high-scale feature map, $X_{K_{2}}$. This asymmetric projection leverages the hierarchical local features of the image to guide the information interaction. The high-level context ($X_{K_{2}}$) acts as the key to unlock the relevant details in the low-level query ($X_{K_{1}}$). The process is mathematically described by the following equations:
\begin{linenomath*}
	\begin{flalign}
		&Q=\mathrm {DW}\left (X_{K_{1}} \right ), K=\mathrm {DW}\left (X_{K_{2}} \right )& \\
		&V_{1}=\mathrm {DW}\left (X_{K_{1}} \right ), V_{2}=\mathrm {DW}\left (X_{K_{2}} \right ) &\\
		&A=\mathrm {Softmax} \left (\frac{QK^{T}}{\sqrt{d_{k}} } \right )&
	\end{flalign}
	\label{equ_3}
\end{linenomath*}
\par\noindent where $A$ represents the attention score matrix, encapsulating the interactive information between scales. To achieve dynamic, adaptive extraction of multi-scale cloud systems, the feature maps derived from the decomposed large-kernel sequence are weighted and fused using these interaction scores. This yields the multi-scale spatially selected feature tensor, $X_{A}$:
\begin{linenomath*}
	\begin{flalign}
		&X_{A_{1}}=A\odot V_{1}, X_{A_{2}}=A\odot V_{2}& \\
		&X_{A}=\mathrm {Cat}\left (X_{A_{1}}, X_{A_{2}} \right ) &
	\end{flalign}
	\label{equ_6}
\end{linenomath*}
\par\noindent where $\odot $ denotes the element-wise multiplication, $\text{Cat}(\cdot)$ denotes the concatenation. This formulation ensures that the fused features $X_{A}$ are not merely a summation of scales, but a context-aware synthesis where global cloud patterns inform the interpretation of local pixels.
\begin{figure}[!t]
	\centering
	\includegraphics[width=3.3in]{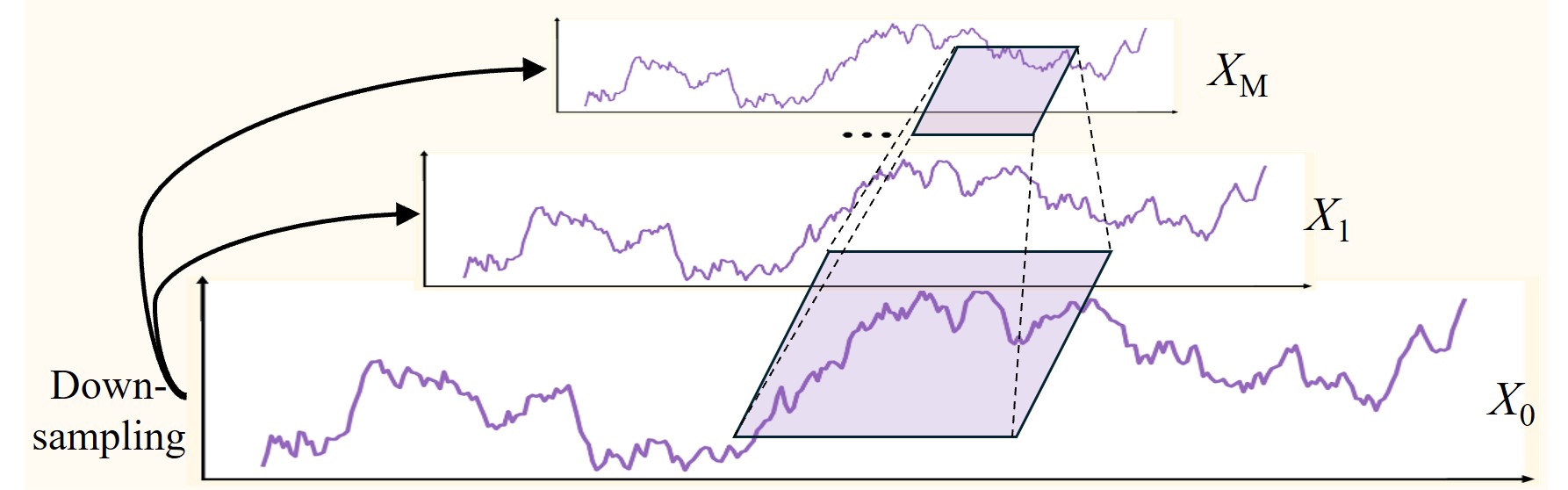}
	\caption{The flowchart of the multi-scale representation.}
	\label{fig_4}
\end{figure}

Recognizing that different spectral channels emphasize distinct image regions (e.g., solar luminance vs. cloud texture), we propose the CE module to dynamically modulate channel importance. As depicted in Fig.~\ref{fig_3}, the fused feature tensor $X_{A}$ is bifurcated into a screening branch (to filter noise) and an enhancement branch (to amplify discriminative features), yielding $X_{CF}$ and $X_{CS}$, respectively. The final output of the CE module is obtained by aggregating the original input $X_{A}$ with the outputs of these two branches. The mechanism incorporates a shared, learnable parameter $\lambda$, which adaptively regulates the information flow:
\begin{linenomath*}
	\begin{flalign}
		&X_{out} = \mathrm{Conv} \left( X_{A} \odot \left(\lambda \cdot \mathrm {Cat}(X_{CS}, X_{CF}) \right )\right) &
	\end{flalign}
	\label{equ_8}
\end{linenomath*}
\par\noindent where $\cdot$ denotes the multiplication. This self-calibrated extraction mechanism effectively suppresses channel redundancy while highlighting critical cloud layer details.

\subsubsection{The Decoder of the MPCS-Net}
The decoder maintains structural consistency with our previous work but reformulates the final output as a context vector to ensure seamless integration with the multimodal fusion process. Specifically, as illustrated on the right side of Fig.~\ref{fig_1}, the decoder employs a recursive upsampling and concatenation strategy to generate multi-scale features $u_{i}$. These features are aggregated into a composite representation, which is processed by a M2B to produce $f_{s}$. Finally, $f_{s}$ is upsampled and concatenated with $u_{1}$ to generate $D_{1}$. The $D_{1}$ undergoes dimensionality reduction via a multi-layer perceptron (MLP), yielding the visual feature mapping $ X_{S}\in \mathbb{R} ^{\left(T-t+1\right )\times d_{model}}$, where $d_{model}$ denotes the dimensionality of the network.
\begin{figure}[!t]
	\centering
	\includegraphics[width=3.3in]{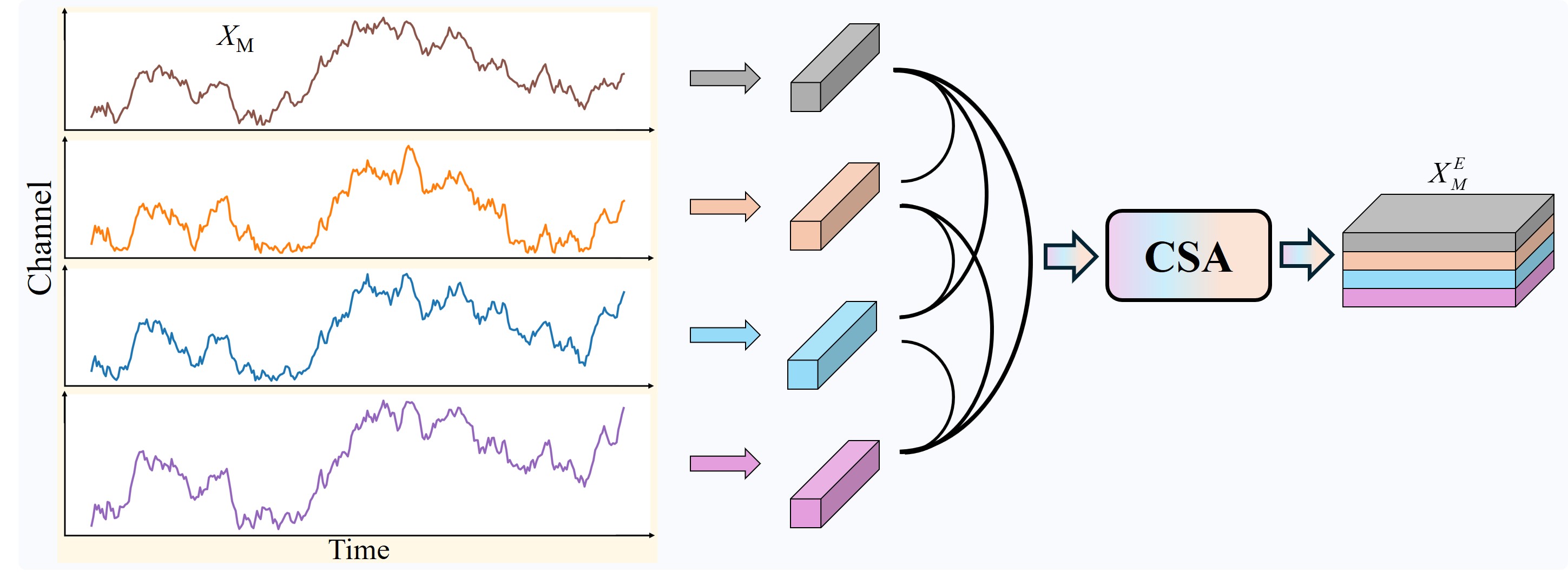}
	\caption{The flowchart of the CSA to capture nonlinear interdependencies among variables on the coarsest scale $X_{M}$.}
	\label{fig_5}
\end{figure}
\subsection{Multi-scale Temporal Imaging Branch Based on SIFR-Net}
PV power generation is governed by complex temporal dynamics characterized by the coupling of long-term climatic trends, daily solar cycles, and transient meteorological fluctuations. Standard Transformer architectures \cite{ViT}, while effective for sequence modeling via self-attention, often struggle to explicitly disentangle these multi-scale periodicities from localized disturbances. This limitation hinders the construction of representation spaces capable of simultaneously capturing global trends and local dynamics. To mitigate this limitation, we introduce SIFR-Net, a multi-scale Sequence to Image analysis network that systematically integrates FFT with a retractable attention mechanism \cite{Retractable}.

\subsubsection{Multi-Scale Pyramid Construction}
The module accepts multivariate time-series data $X_{0} \in \mathbb{R}^{L \times (m+1)}$, comprising historical meteorological parameters and solar irradiance. Inspired by the hierarchical feature maps utilized in Swin Transformers \cite{SViT}, we construct a multi-scale representation $ \mathcal{X}=\left\{X_{0}, X_{1}, \ldots, X_{M}\right\}$ through progressive downsampling, where $X_{m}$ represents the sequence at the $m$-th scale with resolution $L/2^m$, as shown in Fig.~\ref{fig_4}. This multi-scale pyramid architecture enables the model to discern temporal dynamics ranging from fine-grained short-term fluctuations to coarse-grained long-term trends.

To capture nonlinear interdependencies among variables, we implement a channel-wise self-attention (CSA) mechanism at the coarsest scale $X_{M}$. Fig.~\ref{fig_5} illustrates the process. Specifically, given that $X_{M}$ possesses the shortest sequence length, it effectively encapsulates global contextual information, rendering it ideal for modeling cross-variable interactions. This process can be expressed as:
\begin{linenomath*}
	\begin{flalign}
		&X^{E}_{M}=\mathrm {CSA}\left (Q_{M},K_{M},V_{M} \right )&
	\end{flalign}
	\label{equ_9}
\end{linenomath*}
\par\noindent where $Q_{M}$, $K_{M}$ and $V_{M}$ are derived from $X_{M}$ via linear projections. The processed sequences are subsequently projected into a unified feature space via a shared embedding layer:
\begin{linenomath*}
	\begin{flalign}
			&\mathcal{X}_{E}=\left\{X^{E}_{0}, X^{E}_{1}, \ldots, X^{E}_{M}\right\}=Embeding\left (\mathcal{X} \right )&
	\end{flalign}
	\label{equ_10}
\end{linenomath*}
\par\noindent
where $X^{E}_{m} \in \mathbb{R}^ { \left \lceil L/2^m \right \rfloor \times d_{model}}$. This design enables the model to adaptively learn and weight the dependencies among different meteorological and power variables, producing more discriminative fused features and laying a foundation for subsequent periodicity and trend decomposition.
\begin{figure*}[!t]
	\centering{\includegraphics[width=6.5in]{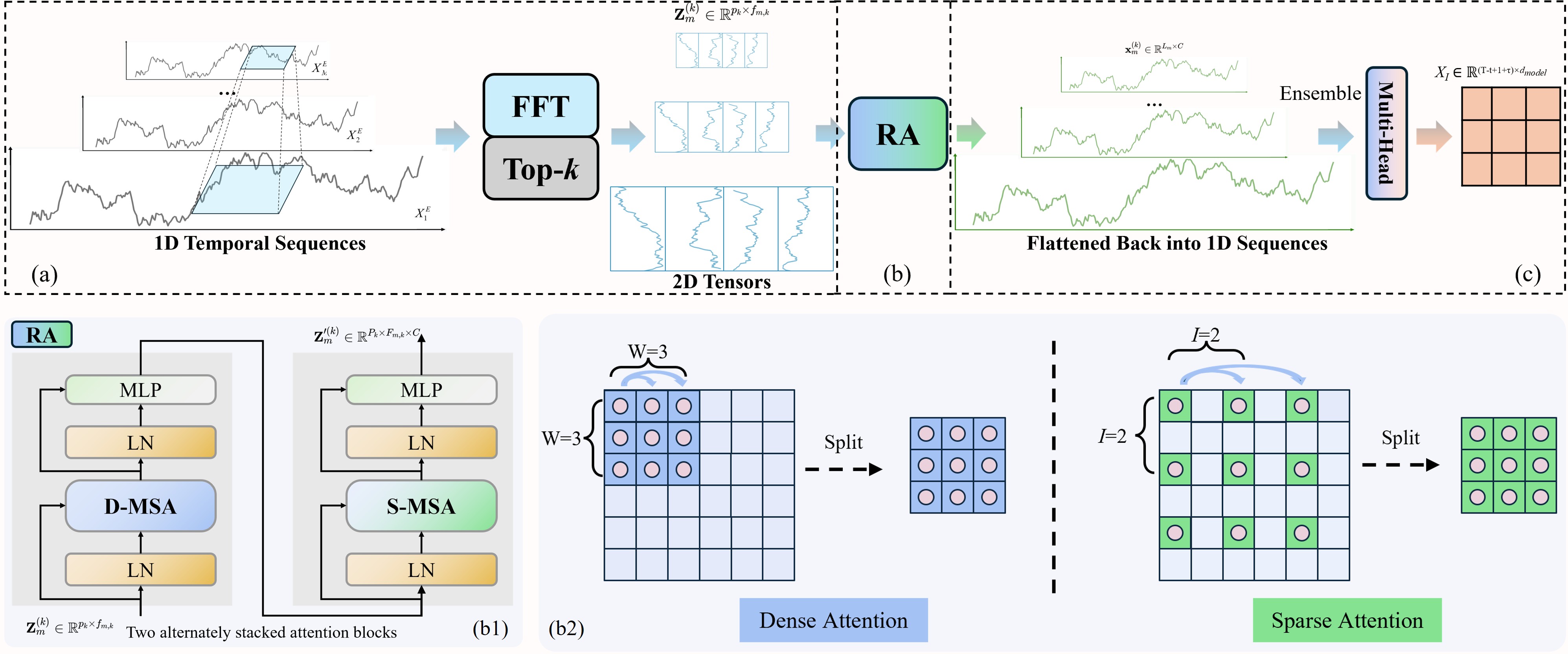}}
	\caption{The flowchart of (a) FFT-based 2D transformation. (b) RA. (c) Amplitude-weighted fusion and prediction. The architecture of (b1) RA, (b2) D-MSA and S-MSA.}
	\label{fig_6}
\end{figure*}
\subsubsection{FFT-based 2D Transformation and Retractable Attention}
A core innovation of SIFR-Net is the transformation of 1D temporal sequences into 2D tensors to leverage image-based attention mechanisms. The sequence and image processing flow is shown in Fig.~\ref{fig_6}. First, an FFT is applied to $X^{E}_{M}$ to identify the top-$k$ dominant frequencies and their corresponding periods $\left \{ p_{1}, p_{2}, \dots , p_{k} \right \}$, generating $k$ two-dimensional tensors accordingly. Based on these periods, sequences at all scales are reshaped into 2D tensors $\mathbf{Z}_{m}^{(k)} \in \mathbb{R}^{p_{k} \times f_{m,k}}$, where the column dimension $ f_{m, k}$ adapts to the sequence length at scale $m$. This restructuring naturally separates intra-period variations (mapped to rows) from inter-period evolution (mapped to columns). In total, the sequence set $\mathcal{X}$ is converted into $\left (m+1 \right) \times k$ multi-resolution images.

Then, to effectively model the superposition of local fluctuations and global trends within these 2D representations, we propose a RA module. While the axial attention mechanism adopted in the literature \cite{TimeMixer} offers one approach to decoupling, its receptive field is relatively fixed. A rigid axial decomposition may fail to adaptively capture cross-scale dependencies that are not strictly aligned with the image rows/columns, or that reside in specific sparse locations. In contrast to rigid axial attention, our method offers a dynamic, “scalable” receptive field through two alternating blocks.

Specifically, as illustrated in Fig.~\ref{fig_6} (b1) and (b2), the module processes the input 2D temporal image $\mathbf{Z}_{m}^{(k)}$ through two alternately stacked attention blocks. The dense attention block (D-MSA) employs local window self-attention. The 2D image is partitioned into non-overlapping $\mathrm {W}   \times \mathrm {W}   $ windows, and standard self-attention is computed within each window. This allows the model to focus on relationships among time points within local neighborhoods, capturing high-frequency, short-period seasonal patterns as well as continuous local textures in the image. The computation is formulated as follows:
\begin{linenomath*}
	\begin{flalign}
		&\mathbf{Q}^{(j)}, \mathbf{K}^{(j)}, \mathbf{V}^{(j)} =\mathbf{Z}^{(j)} \mathbf{W}_{Q}, \mathbf{Z}^{(j)} \mathbf{W}_{K}, \mathbf{Z}^{(j)} \mathbf{W}_{V}& \\
		&\text { D-MSA }\left(\mathbf{Z}^{(j)}\right) =\operatorname{Softmax}\left(\frac{\mathbf{Q}^{(j)}\left(\mathbf{K}^{(j)}\right)^{T}}{\sqrt{d_{k}}}\right) \mathbf{V}^{(j)}&
		\label{equ_11}
	\end{flalign}
\end{linenomath*}
\par\noindent where $\mathbf{Z}^{(j)} \in \mathbb{R} ^ {W^{2} \times C}$ within the $j$-th window, $W_{*}$ represents learnable projection weights. All window computations are performed in parallel, and the output features are subsequently reassembled into the original image layout. 

The sparse attention block (S-MSA) employs a grid-based sparse sampling strategy. Pixels are sampled at fixed intervals $I$, and global attention is computed among these sparse points. This allows any time point in the image to interact with distant yet potentially semantically relevant points elsewhere. This enables the model to capture long-range dependencies and slowly evolving trends that span across multiple periods. The computation proceeds as follows:
\begin{linenomath*}
	\begin{flalign}
	&\mathbf{Z}_{\text {sparse }} =\operatorname{Gather}(\mathbf{Z}, I)& \\
	&\mathbf{Q}_{s}, \mathbf{K}_{s}, \mathbf{V}_{s} =\mathbf{Z}_{\text {sparse }} \mathbf{W}_{Q}^{s}, \mathbf{Z}_{\text {sparse }} \mathbf{W}_{K}^{s}, \mathbf{Z}_{\text {sparse }} \mathbf{W}_{V}^{s} &\\
	&\operatorname{S-MSA}(\mathbf{Z}) =\operatorname{Scatter}\left(\operatorname{Softmax}\left(\frac{\mathbf{Q}_{s} \mathbf{K}_{s}^{T}}{\sqrt{d_{k}}}\right) \mathbf{V}_{s}, I\right)&
	\end{flalign}
	\label{equ_13}
\end{linenomath*}
\par\noindent where the $\mathrm {Gather}\left (\cdot  \right )$ operation samples features along the spatial dimension at interval $I$, while $\mathrm{Scatter}\left (\cdot  \right )$ performs the inverse, interpolating the updated sparse features back to their original positions. $I$ is a hyperparameter that controls the “sparsity” and globality of the receptive field. A larger $I$ results in fewer sampled points participating in the computation but enables each point to interact directly with features from more distant locations.

\subsubsection{Amplitude-Weighted Fusion and Prediction}
After processing by the retractable attention module, a refined 2D feature map $\mathbf{Z}_{m}^{\prime(k)} \in \mathbb{R}^{P_{k}\times F_{m,k}\times C}$ is obtained for each scale $m$ and each period $k$. Then the 2D feature maps are flattened back into 1D sequences. We perform a reshaping operation inverse to the imaging process:
\begin{linenomath*}
	\begin{flalign}
		&\mathbf{x}_{m}^{(k)}=\operatorname{Reshape}_{2 D \rightarrow 1 D}\left(\mathbf{Z}_{m}^{\prime(k)}\right) \in \mathbb{R}^{L_{m} \times C}&
	\end{flalign}
	\label{equ_16}
\end{linenomath*}
\par\noindent where $L_{m}=P_{k}\times F_{m,k}$ equals the original sequence length at scale $m$, thereby ensuring temporal alignment. To synthesize the information extracted from different periodic views, we employ an amplitude-weighted fusion strategy. Specifically, using the amplitude spectrum $\mathbf{A}=\{A_{f_{1}},\dots ,A_{f_{K}} \}$ extracted during the imaging phase from the coarsest scale, which reflects the relative importance of each period, normalized weights are computed as:
\begin{linenomath*}
	\begin{flalign}
		&\left\{\hat{A}_{f_{k}}\right\}_{k=1}^{K}=\operatorname{Softmax}\left(\left\{A_{f_{k}}\right\}_{k=1}^{K}\right)&
	\end{flalign}
	\label{equ_17}
\end{linenomath*}
\par\noindent then the fused feature $\mathbf{h}_{m}$ at scale $m$ is given by:
\begin{linenomath*}
	\begin{flalign}
		&\mathbf{h}_{m}=\sum_{k=1}^{K} \hat{A}_{f_{k}} \cdot \mathbf{x}_{m}^{(k)}, \quad m \in\{0, \cdots, M\}&
	\end{flalign}
	\label{equ_18}
\end{linenomath*}
\par\noindent This step adaptively enhances the expression of dominant periodic patterns in the features.

Finally, a multi-scale prediction head generates an ensemble forecast $X_{I} \in \mathbb{R}^{(T-t+1+\tau) \times d_{\text{model}}}$, integrating contributions from all scales to ensure robustness against both high-frequency noise and long-term drift.
\begin{linenomath*}
	\begin{flalign}
		&X_{I}=\operatorname{Ensemble}\left(\operatorname{Head}_{0}\left(\mathbf{h}_{0}\right), \ldots, \operatorname{Head}_{M}\left(\mathbf{h}_{M}\right)\right)&
	\end{flalign}
	\label{equ_19}
\end{linenomath*}

\subsection{Multi-model Fusion Branch Based on Mamba}
The preceding branches provide two distinct representations of the PV system state: spatial features capturing cloud topology ($ X_{S}\in \mathbb{R} ^{\left(T-t+1\right )\times d_{model}}$) and temporal features encoding periodic power dynamics ($ X_{I}\in \mathbb{R} ^{\left(T-t+1+\tau \right )\times d_{model}}$). Conventional fusion techniques, such as simple concatenation or linear superposition, often fail to model the dynamic, time-varying interactions between these modalities, particularly over long sequences where dependencies may shift rapidly. To overcome this limitation, we introduce the MMIF module, underpinned by the Mamba architecture.
\begin{figure}[!t]
	\centering{\includegraphics[width=3.3in]{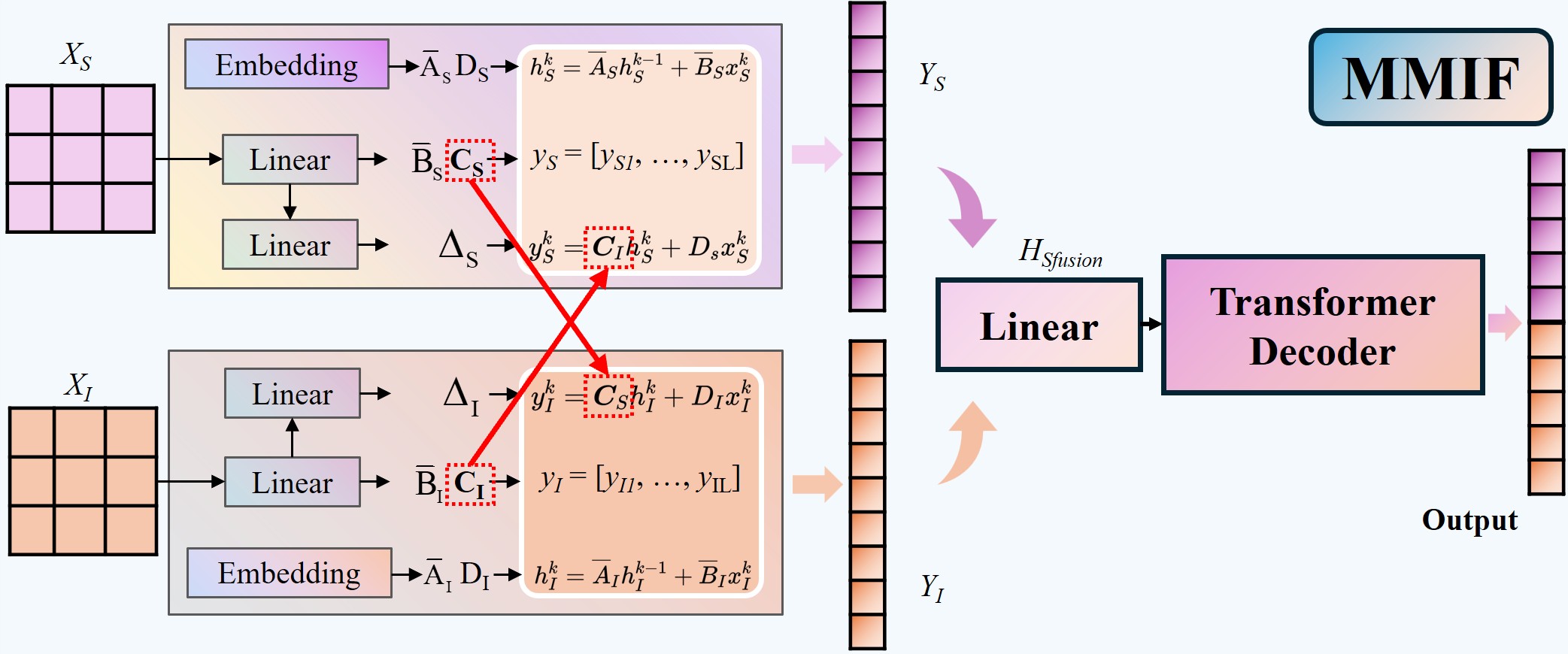}}
	\caption{The architecture of the proposed MMIF.}
	\label{fig_7}
\end{figure}
\subsubsection{Multimodal Interaction Fusion}
As a selective state-space model \cite{Vmamba}, Mamba offers the advantages of linear complexity and global receptive fields when processing long sequences. Its core mechanism lies in the dynamic generation of model parameters (e.g., $\overline{B}$, $\overline{C}$, and $\Delta $) based on the input, enabling selective retention and propagation of critical information. To facilitate deep interaction between the two modalities, we introduce a cross-modal “C-matrix swapping” mechanism. While recent works like Sigma and STT \cite{Sigma, STT} have explored parameter exchange in Mamba for static multimodal image segmentation, M3S-Net uniquely adapts this mechanism to the spatiotemporal domain. Unlike Sigma and STT, which align spatially registered modalities (e.g., RGB-Thermal), our approach aligns high-frequency temporal dynamics with spatial cloud features, addressing the specific challenge of time-lagged causality in PV forecasting.

Specifically, as illustrated in Fig.~\ref{fig_7}, the sequence of each modality is fed into a Mamba block that shares the same basic structure. When computing the output, we swap the $\overline{C}$ matrices (the state-to-output mapping matrices) generated by each modality. As a result, each modality produces its output by relying not only on its own historical state but also on contextual information embedded in the state of the other modality. This process can be formally expressed as follows:
\begin{linenomath*}
	\begin{flalign}
	&h_{S}^{k} =\overline{A}_{S} h_{S}^{k-1}+\overline{B}_{S}x_{S}^{k}&\\
	&y_{S}^{k} =\boldsymbol{C}_{I} h_{S}^{k}+D_{s} x_{S}^{k}&\\
	&h_{I}^{k} =\overline{A}_{I} h_{I}^{k-1}+\overline{B}_{I}x_{I}^{k} &\\
	&y_{I}^{k} =\boldsymbol{C}_{S} h_{I}^{k}+D_{I} x_{I}^{k}&
	\end{flalign}
	\label{equ_20}
\end{linenomath*}
\par\noindent where $k$ denotes the sequence position index, $h$ represents the hidden state, and $\overline{A}$, $\overline{B}$, $D$ are the parameterized matrices specific to each modality. The key step lies in the fact that the output $ y_{s}^{k}$ of the ground-based cloud image modality $X_{S}$ utilizes the mapping matrix $\boldsymbol{C}$ generated by the temporal modality $X_{I}$, and vice versa. This design forces the model to integrate the feature patterns extracted by the other modality when decoding the current information of each modality. Consequently, tight cross-modal correlations are established across all positions of the sequence, enabling genuine end-to-end information interaction.
\subsubsection{Feature Fusion and Decoding Prediction}
Following the cross-modal interaction, the mutually informed output sequences $Y_s$ and $Y_I$ are concatenated and projected via a linear layer to form the unified fusion feature $H_{fusion}$. A lightweight Transformer decoder then autoregressively generates the final PV power forecast. This decoder captures the dependencies within the fused representation, mapping the synthesized cross-modal context to the precise power output values for the target horizon.
\begin{linenomath*}
	\begin{flalign}
	&H_{fusion }=\operatorname{Linear}\left(\operatorname{Cat}\left(Y_{ s }, Y_{ I }\right)\right)&
	\end{flalign}
	\label{equ_24}
\end{linenomath*}

\section{Experimental Setting}\label{sec:setting}
The experimental framework was established using Python 3.9 within the PyTorch 1.10.1 ecosystem, providing a robust environment for DL deployment. To accommodate the computational intensity of multimodal feature fusion, all training and evaluation procedures were executed on a high-performance server running Ubuntu 21.04. The hardware infrastructure consisted of an Intel(R) Xeon(R) Gold 5218N CPU clocked at 2.3 GHz and an NVIDIA GeForce RTX 4090 GPU, equipped with 24 GB of VRAM to support large-scale tensor operations. The remainder of this section contains the dataset, evaluation metrics, and baseline models.
\begin{table}[!t]
	\centering
	\caption{Cloud classification standard for CSRC dataset.}
	\resizebox{\linewidth}{!}{
		\begin{tabular}{ccc}
			\toprule
			Cloud class & Shape and characteristic & Color \\
			\midrule
			White cloud & \makecell{White transparent clouds, \\ such as cirrus and cirrocumulus, \\ are pure white and translucent \\ or appear as white scaly flakes.} & White \\
			Gray cloud & \makecell{Gray clouds, \\ such as stratus and altostratus, \\ are gray and foggy with a dark base.} & Gray \\
			Sun & Radiation source & Red \\
			Background & There are no clouds in the sky & Blue \\
			\bottomrule
	\end{tabular}}
	\label{Table1}
\end{table}
\subsection{Dataset}\label{sec:data}
PV power forecasting accuracy is inextricably linked to the fidelity and informational density of the input variables. Singlemodal time-series data inevitably prove insufficient for characterizing the direct modulation of surface solar irradiance by dynamic sky conditions, particularly those driven by stochastic cloud kinematics. In the context of ultra-short-term forecasting, rapid cloud transients, encompassing morphological evolution and abrupt displacement, induce sharp fluctuations in irradiance, which subsequently destabilize PV power output. Consequently, the integration of ground-based cloud imagery, which captures localized atmospheric states with high spatiotemporal resolution, is paramount for enhancing predictive precision.

Existing multimodal datasets for PV forecasting frequently suffer from a critical limitation: ground-based cloud imagery is often reduced to binary masks distinguishing only between cloud and sky, thereby neglecting critical attributes such as cloud scale, morphology, and color features. Such coarse-grained representations are inadequate for satisfying the stringent requirements of ultra-short-term forecasting models, which demand high-fidelity environmental data.

Addressing this deficiency, our prior research introduced a novel ground-based cloud image dataset that incorporates complex-scale attributes along with radiation source and color characteristics \cite{MPCM-Net}. This dataset emphasizes cloud scale and color properties and introduces fine-grained segmentation of clouds based on radiation sources, stratifying datasets into four distinct classes (details in Table~\ref{Table1}). 

The observational infrastructure was deployed in Xiqing District, Tianjin, China (Geographic coordinates: 117.03° E, 39.10° N). Visual data were acquired using an All-Sky Imager (ASI-DC-TK02), encased in a weather-resistant enclosure and offering a field of view exceeding 180°. The device was calibrated to capture sky images at fixed 30-second intervals. All imagery is archived in RGB JPEG format with a resolution of 1260 $\times$1260 pixels. To ensure ground truth reliability, fine-grained annotations were generated by trained meteorologists utilizing LabelMe software, with subsequent cross-validation performed to guarantee label consistency. This dataset facilitates the pixel-wise quantification of sunlight attenuation by varying cloud types, serving as a foundational resource for mapping multimodal inputs to PV generation.
\begin{figure*}[!t]
	\centering{\includegraphics[width=6in]{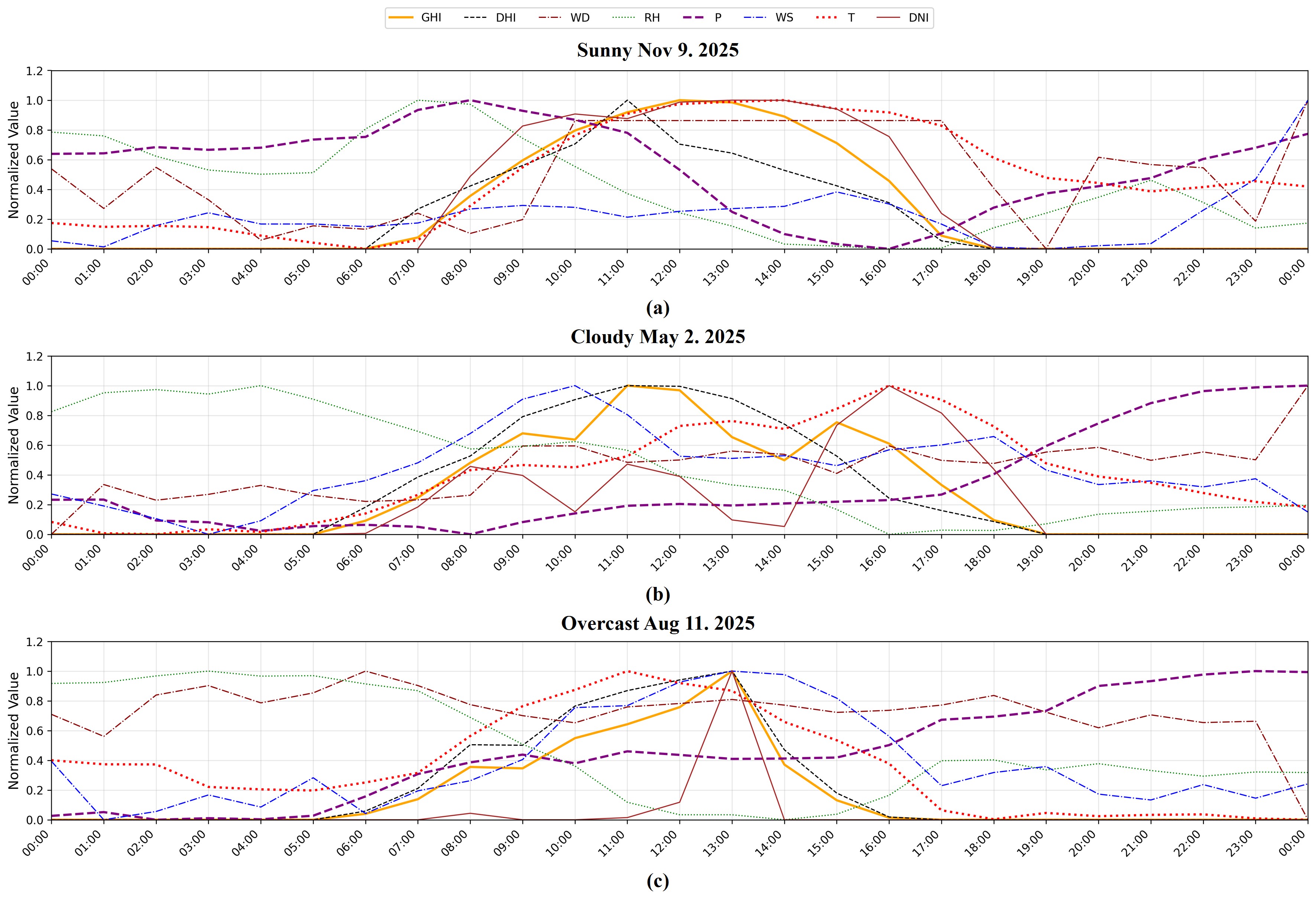}}
	\caption{Eight types of meteorological data under three types of typical weather.}
	\label{fig_8}
\end{figure*}

Complementing the visual data, our proposed Fine-Grained PV Power Dataset integrates high-frequency meteorological records with PV power output. Meteorological parameters were logged by an on-site weather monitoring station at 10-minute intervals, synchronized with the imaging frequency. The monitored variables include: GHI ($\mathrm {W/m^2}$), DNI ($\mathrm {W/m^2}$), diffuse horizontal irradiance (DHI, $\mathrm {W/m^2}$), wind speed (WS, $\mathrm {m/s}$), wind direction (WD, $\mathrm {^\circ}$), temperature (T, $\mathrm {^\circ C}$), relative humidity (RH), and atmospheric pressure (P, $\mathrm {Pa}$). Data collection spanned the full calendar year of 2025 (January 1 to December 31), encompassing a diverse array of meteorological conditions, including sunny, cloudy, and overcast, to ensure temporal representativeness. Fig.~\ref{fig_8} elucidates the synchronous fluctuations of these meteorological data streams on a representative day, visually corroborating the coupling between cloud dynamics and irradiance attenuation.

The aggregate dataset encompasses 59 sunny days, 191 cloudy days, and 50 overcast days. To optimize training efficiency and model generalization, a selective filtration strategy excluded the majority of monotonic sunny and stable overcast data points, which contain limited informational variance. Conversely, a higher proportion of partly cloudy data was retained. This strategy effectively reduces dataset volume while compelling the deep learning model to learn robust features associated with complex cloud interactions. Post-alignment of imagery and meteorological sequences, 41,780 valid data points were identified. These were randomly shuffled and stratified into training, validation, and test sets in a 7:1:2 ratio.

\subsection{Evaluation Metrics}\label{sec:eva}
To facilitate a rigorous assessment of the M3S-Net model's efficacy in PV power forecasting, this study adopts a comprehensive suite of evaluation metrics spanning point-prediction accuracy, trend consistency, and segmentation quality.

Fundamentally, PV power forecasting constitutes a regression problem necessitating measuring the deviation between predicted and observed values. This paper employs the following metrics: MAE, MSE, normalized root mean squared error (NRMSE), and $\mathrm { R^{2}}$. MAE quantifies the mean magnitude of errors without considering their direction, offering intuitive interpretability. MSE heavily penalizes large deviations, serving as a critical indicator for ramp event prediction failure. NRMSE facilitates performance comparison across disparate scales by normalizing the error against the data range. $\mathrm { R^{2}}$ indicates the proportion of variance in the dependent variable explained by the model; values approaching 1 denote superior goodness-of-fit. The formulas for these metrics are given below:
\begin{linenomath*}
	\begin{flalign}
		&\mathrm {MAE}=\frac{1}{N}\sum_{i=1}^{N}  \left | y_{i}-\hat{y}_{i}  \right | &\\
		&\mathrm {MSE}=\frac{1}{N}\sum_{i=1}^{N}  \left ( y_{i}-\hat{y}_{i}  \right )^{2} &\\
		&	\mathrm {NRMSE}=\frac{1}{\max (y)-\min (y)} \sqrt{\frac{1}{N} \sum_{i=1}^{N}\left(y_{i}-\hat{y}_{i}\right)^{2}} \times 100 \%  &\\
		&\mathrm { R^{2}} =1-\frac{\sum_{i=1}^{N}\left(y_{i}-\hat{y}_{i}\right)^{2}}{\sum_{i=1}^{N}\left(y_{i}-\bar{y}\right)^{2}}&
	\end{flalign}
	\label{equ_25}
\end{linenomath*}
\par\noindent where $ y_{i}$ is the true value, $\hat{y}_{i}$ is the predicted value, $\bar{y}$ is the mean of the true values, and $N$ denotes the sample size.

For ground based cloud image segmentation, standard image segmentation metrics are used, including Precision (P), Recall (R), and Mean Intersection over Union (MIoU). The formulas are:
\begin{linenomath*}
	\begin{flalign}
		&\mathrm { P} =\frac{TP}{TP+FP} &\\
		&\mathrm { R}=\frac{TP}{TP+FN} &\\
		&\mathrm { MIoU} =\frac{1}{C}\sum_{c=1}^{C}\frac{TP_{c}}{TP_{c}+FP_{C}+FN_{c}}&
	\end{flalign}
	\label{equ_29}
\end{linenomath*}
\par\noindent where $TP$ denotes true positives, $FP$ denotes false positives, $FN$ denotes false positives, and $C$ denotes the number of classes.

During network training, MSE serves as the loss function for the PV forecasting task due to its continuous differentiability and efficacy in gradient descent optimization. Concurrently, binary cross-entropy (BCE) loss governs the optimization of the pixel-wise cloud classification branch.
\begin{table}[!t]
	\centering
	\caption{Quantitative comparison with different methods on FGPD for GCI segmentation. $\downarrow$ (or $\uparrow$) indicates lower (or higher) is better. The best results are highlighted in bold.}
	{\small
		\begin{tabular}{cccc}
			\toprule
			Method & P($\uparrow $) & R($\uparrow $) & MIoU($\uparrow $) \\
			\midrule
			U-Net & 66.7 & 64.6 & 46.7\\
			SegFormer & 84.6 & 85.2 & 61.3\\
			CloudFU-Net & 86.9 & 86.5 & 62.1\\
			CloudSwinNet & 87.1 & 87.6 & 62.3\\
			MPCM-Net & 90.6 & 91.1 & 64.8\\
			MPCS-Net & \textbf{91.3} & \textbf{91.8} & \textbf{65.6}\\
			\bottomrule
	\end{tabular}}
	\label{Table2}
\end{table}
\begin{table*}[ht!]
	\centering
	\caption{Comparison results of GHI based on MAE ($\mathrm {W/m^2}$) on the FGPD.}
	{\small
		\begin{tabular}{cccccccc}
			\toprule
			\multirow{2}{*}{Model structure} & \multirow{2}{*}{Method} & \multicolumn{6}{c}{MAE ($\mathrm {W/m^2}$) ($\downarrow$)} \\ 
			\cmidrule(lr){3-8}
			& & 10 MA & 20 MA & 30 MA & 40 MA & 50 MA & 60 MA \\
			\midrule
			
			\multirow{3}{*}{Singlemodal} 
			& LSTM     & 39.76 & 42.28 & 45.53 & 48.97 & 54.64 & 61.42 \\
			& DLinear  & 22.32 & 31.44 & 44.33 & 55.62 & 66.73 & 77.64 \\
			& TimesNet & 23.45 & 28.61 & 34.25 & 41.27 & 47.82 & 55.66 \\
			
			\multirow{4}{*}{Multimodal} 
			& CNN-based   & 22.56 & 27.74 & 33.96 & 40.66 & 47.12 & 54.12 \\
			& ViT-based      & 22.37 & 28.13 & 33.84 & 39.72 & 47.33 & 53.83 \\
			& T2T & 21.16 & 26.64 & 32.17 & 38.43 & 45.67 & 53.12 \\
			& M3S-Net & \textbf{19.84} & \textbf{24.43} & \textbf{30.76} & \textbf{37.14} & \textbf{43.51} & \textbf{50.35} \\
			\bottomrule
	\end{tabular}}
	\label{Table3}
\end{table*}
\begin{table*}[ht!]
	\centering
	\caption{Comparison results of GHI based on NRMSE ($\%$) on the FGPD.}
	{\small
		\begin{tabular}{cccccccc}
			\toprule
			\multirow{2}{*}{Model structure} & \multirow{2}{*}{Method} & \multicolumn{6}{c}{NRMSE ($\%$) ($\downarrow$)} \\ 
			\cmidrule(lr){3-8}
			& & 10 MA & 20 MA & 30 MA & 40 MA & 50 MA & 60 MA \\
			\midrule
			
			\multirow{3}{*}{Singlemodal} 
			& LSTM     & 6.82 & 7.31 & 7.94 & 8.82 & 9.41 & 10.23 \\
			& DLinear  & 5.14 & 5.72 & 7.16 & 8.73 & 10.26 & 11.78 \\
			& TimesNet & 5.42 & 5.96 & 6.78 & 7.64 & 8.86 & 9.35 \\
			
			\multirow{4}{*}{Multimodal} 
			& CNN-based   & 5.23 & 5.88 & 6.52 & 7.36 & 8.29 & 9.14 \\
			& ViT-based   & 5.09 & 5.63 & 6.38 & 7.14 & 7.93 & 8.89 \\
			& T2T & 4.84 & 5.36 & 6.03 & 6.85 & 7.72 & 8.68 \\
			& M3S-Net & \textbf{4.61} & \textbf{5.14} & \textbf{5.78} & \textbf{6.56} & \textbf{7.42} & \textbf{8.51} \\
			\bottomrule
	\end{tabular}}
	\label{Table4}
\end{table*}

\begin{table}[!t]
	\centering
	\caption{Comparison results of GHI based on $\mathrm { R^{2}}$  on the FGPD.}
	\resizebox{\linewidth}{!}{
		\begin{tabular}{cccccccc}
			\toprule
			\multirow{2}{*}{\makecell[c]{Model \\structure}} & \multirow{2}{*}{Method} & \multicolumn{6}{c}{$\mathrm { R^{2}}$ ($\uparrow$)} \\ 
			\cmidrule(lr){3-8}
			& & \makecell[c]{10 \\MA} & \makecell[c]{20 \\MA} & \makecell[c]{30 \\MA} & \makecell[c]{40 \\MA} & \makecell[c]{50 \\MA} & \makecell[c]{60 \\MA} \\
			\midrule
			
			\multirow{3}{*}{\makecell[c]{Single\\-modal}} 
			& LSTM     & 0.912 & 0.903 & 0.894 & 0.882 & 0.864 & 0.845 \\
			& DLinear  & 0.931 & 0.922 & 0.910 & 0.897 & 0.872 & 0.841 \\
			& TimesNet & 0.947 & 0.923 & 0.904 & 0.884 & 0.849 & 0.807 \\
			
			\multirow{4}{*}{\makecell[c]{Multi\\-modal}} 
			& CNN-based   & 0.954 & 0.931 & 0.916 & 0.902 & 0.874 & 0.851 \\
			& ViT-based   & 0.953 & 0.934 & 0.915 & 0.894 & 0.871 & 0.848 \\
			& T2T & 0.958 & 0.945 & 0.928 & 0.902 & 0.885 & 0.864 \\
			& M3S-Net & \textbf{0.966} & \textbf{0.947} & \textbf{0.931} & \textbf{0.911} & \textbf{0.893} & \textbf{0.872} \\
			\bottomrule
	\end{tabular}}
	\label{Table5}
\end{table}

\subsection{Benchmark Models}
To benchmark the performance of the proposed M3S-Net, a diverse array of SOTA models was selected for comparison. The fine-grained visual extraction branch is evaluated against both classical semantic segmentation architectures and specialized cloud segmentation networks. These include CNN-based models such as U-Net \cite{U-Net}, CloudFU-Net \cite{CloudFU-Net}, and MPCM-Net \cite{MPCM-Net}, alongside Vision Transformer (ViT) architectures like SegFormer \cite{SegFormer} and CloudSwinNet \cite{CloudSwinNet}. The forecasting capabilities are benchmarked against established time-series models (LSTM \cite{LSTM}, DLinear \cite{DLinear}, and TimesNet \cite{Timesnet}) and multimodal fusion frameworks, including CNN-based \cite{CNN-based}, ViT-based \cite{ViT-based}, and T2T \cite{T2T-ViT}. All experimental results cited in this study were derived either from open-source implementations or directly from the original authors' publications to ensure a fair and reproducible comparative analysis.

\section{Experimental Results and Discussions}\label{sec:exp}
To rigorously evaluate the efficacy of the proposed M3S-Net, a comprehensive comparative analysis was conducted against established baseline models for ultra-short-term PV power forecasting. The prediction horizons were stratified from 10 to 60 minutes-ahead (MA), corresponding to 1 to 6 time steps, to assess model performance across varying temporal scales.

\subsection{Overall Performance}
The quantitative evaluation results of the proposed MPCS-Net and other segmentation methods on GCI of our FPGD. Table~\ref{Table2} demonstrates the superior performance of our proposed MPCS-Net over all comparative GCI segmentation methods on FGPD. Specifically, MPCS-Net achieves MIoU improvements of 0.8$\%$, compared to the previous MPCM-Net \cite{MPCM-Net}. The performance advantage stems from our method, from its CSIA within SCSM, which enhances inter-scale interaction within the encoder, coupled with the CE module that emphasizes critical inter-regional features.

The comparative results between the proposed method and the benchmark models are presented in Tables~\ref{Table3} through ~\ref{Table5}. M3S-Net consistently demonstrated superior performance across all prediction horizons. Specifically, within the critical 10-minute forecast window, M3S-Net achieved a significant reduction in MAE and NRMSE of approximately 6.2$\%$ and 4.8$\%$, respectively, relative to the second-best performing model. Furthermore, the model attained a $\mathrm { R^{2}}$ of 0.964, indicating exceptional predictive precision. As the prediction horizon extended to 60 minutes, M3S-Net exhibited the slowest rate of performance degradation among all tested models, which corroborates the model’s enhanced robustness and stability in longer-term forecasting scenarios. 

These comparative experiments elucidate the distinct advantages inherent in fusing multi-source data. Models relying exclusively on time-series data, such as LSTM, DLinear, and TimesNet, performed adequately under stable meteorological conditions. However, during periods of rapid irradiance fluctuation induced by cloud cover, these singleimodal models exhibited significantly increased prediction errors and pronounced phase lag. In contrast, models incorporating ground-based cloud imaging (e.g., CNN-based, ViT-based, and T2T) demonstrated superior overall performance. This disparity validates the hypothesis that spatial distribution and dynamic information derived from cloud images are indispensable for capturing the underlying causality of power fluctuations.
\begin{figure}[!t]
	\centering{\includegraphics[width=3in]{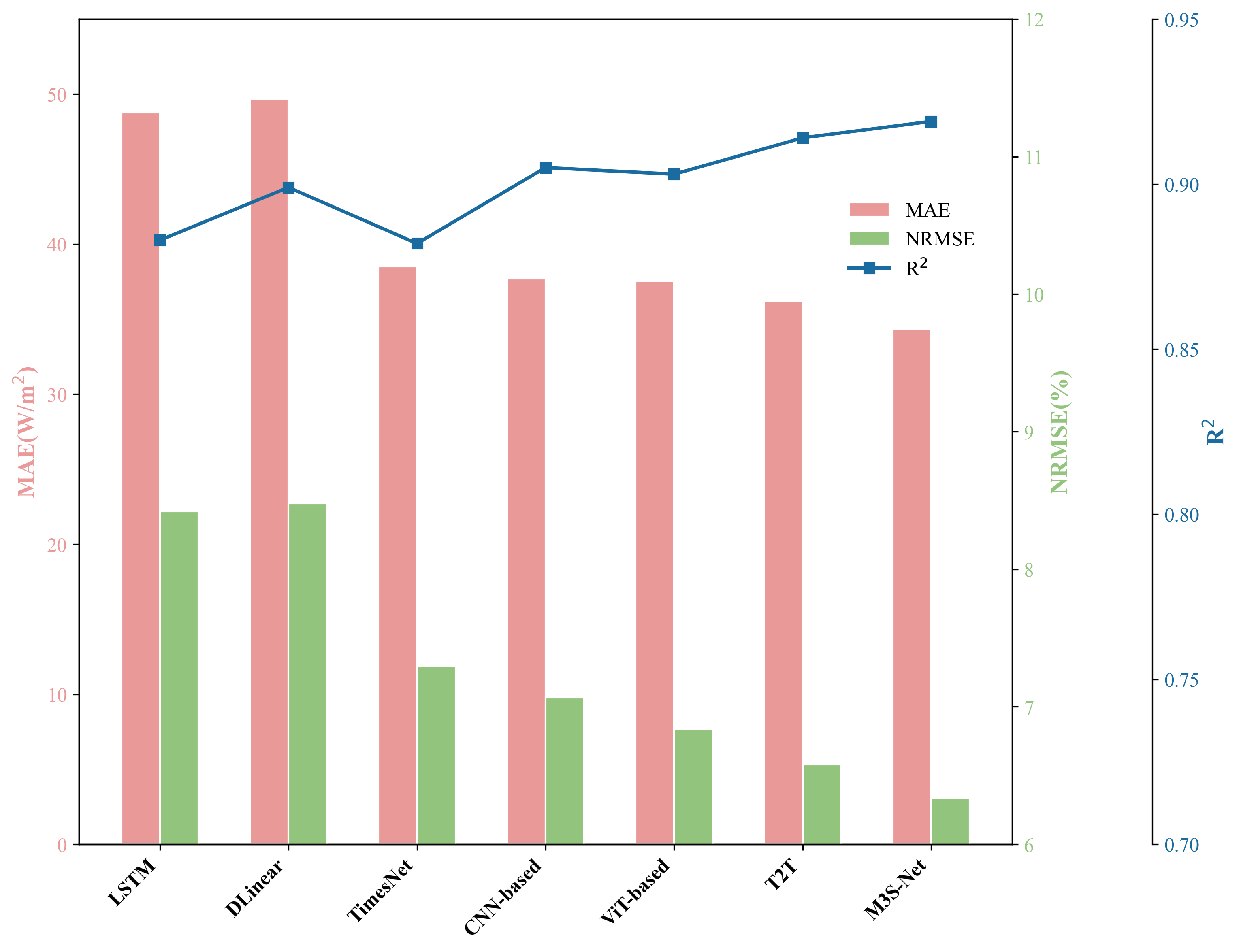}}
	\caption{Visual results of average value of different error indicators.}
	\label{fig_9}
\end{figure}

M3S-Net significantly outperformed other multimodal fusion architectures, a success attributable to the synergistic engineering of its three core branches. First, the MPCS-Net facilitates fine-grained extraction of multi-scale cloud features and, crucially, the resolution of semi-transparent cloud boundaries. Second, the SIFR-Net explicitly models the multi-scale periodicity and trends inherent in power data via time-series imaging and scalable attention mechanisms. Third, the MMIF module enables deep feature complementarity and information exchange that far exceeds the capabilities of simple concatenation or linear fusion strategies. Additionally, the bar chart in Fig.~\ref{fig_9}, which compares the average errors across the 10-to-60-minute horizon, visually underscores that M3S-Net maintains the lowest error magnitude.
\begin{figure*}[!t]
	\centering{\includegraphics[width=6in]{ 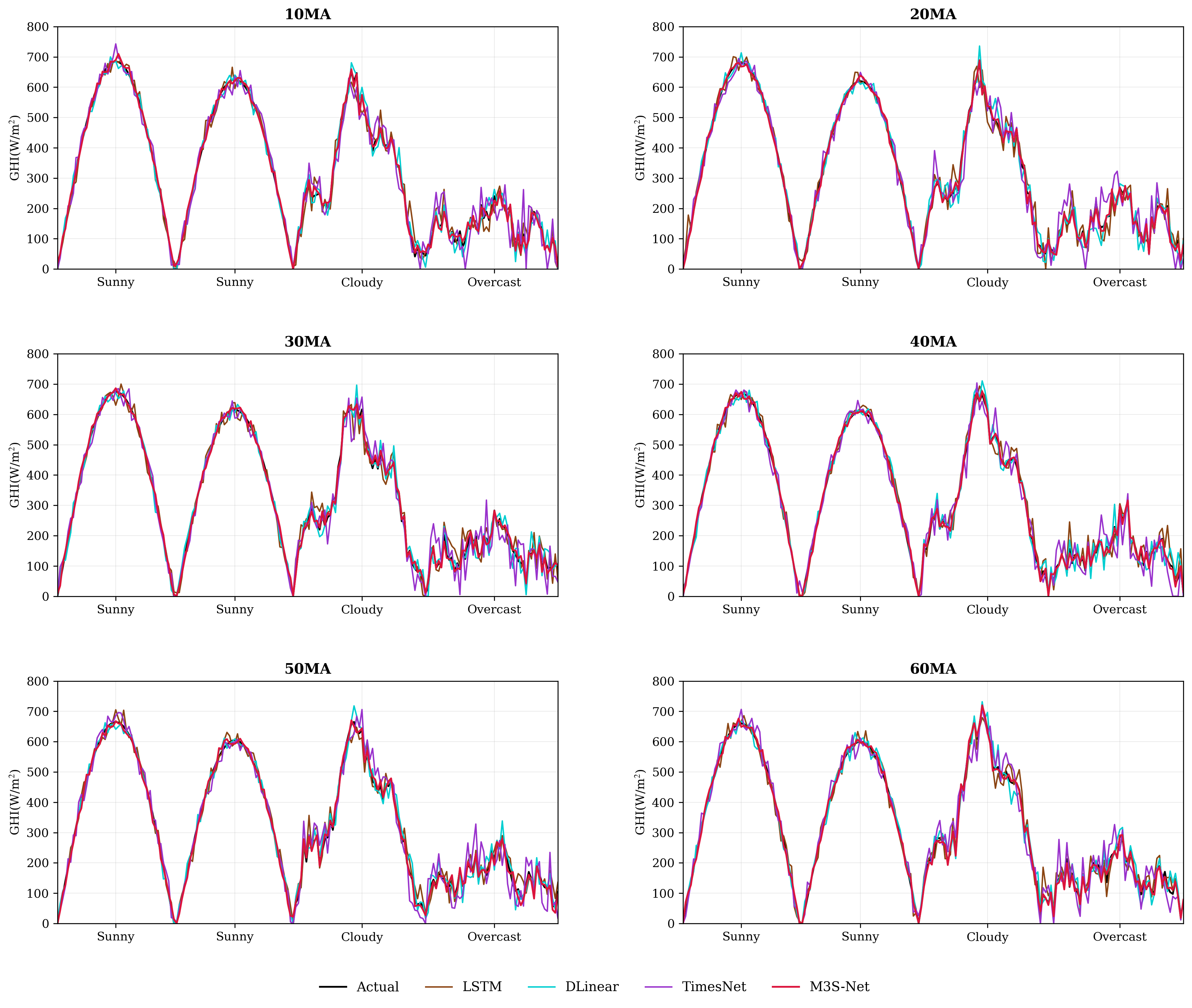}}
	\caption{ Visual results under three typical weather categories of the proposed M3S-Net and other singlemodal methods.}
	\label{fig_10}
\end{figure*}
\begin{figure*}[!t]
	\centering{\includegraphics[width=6in]{ 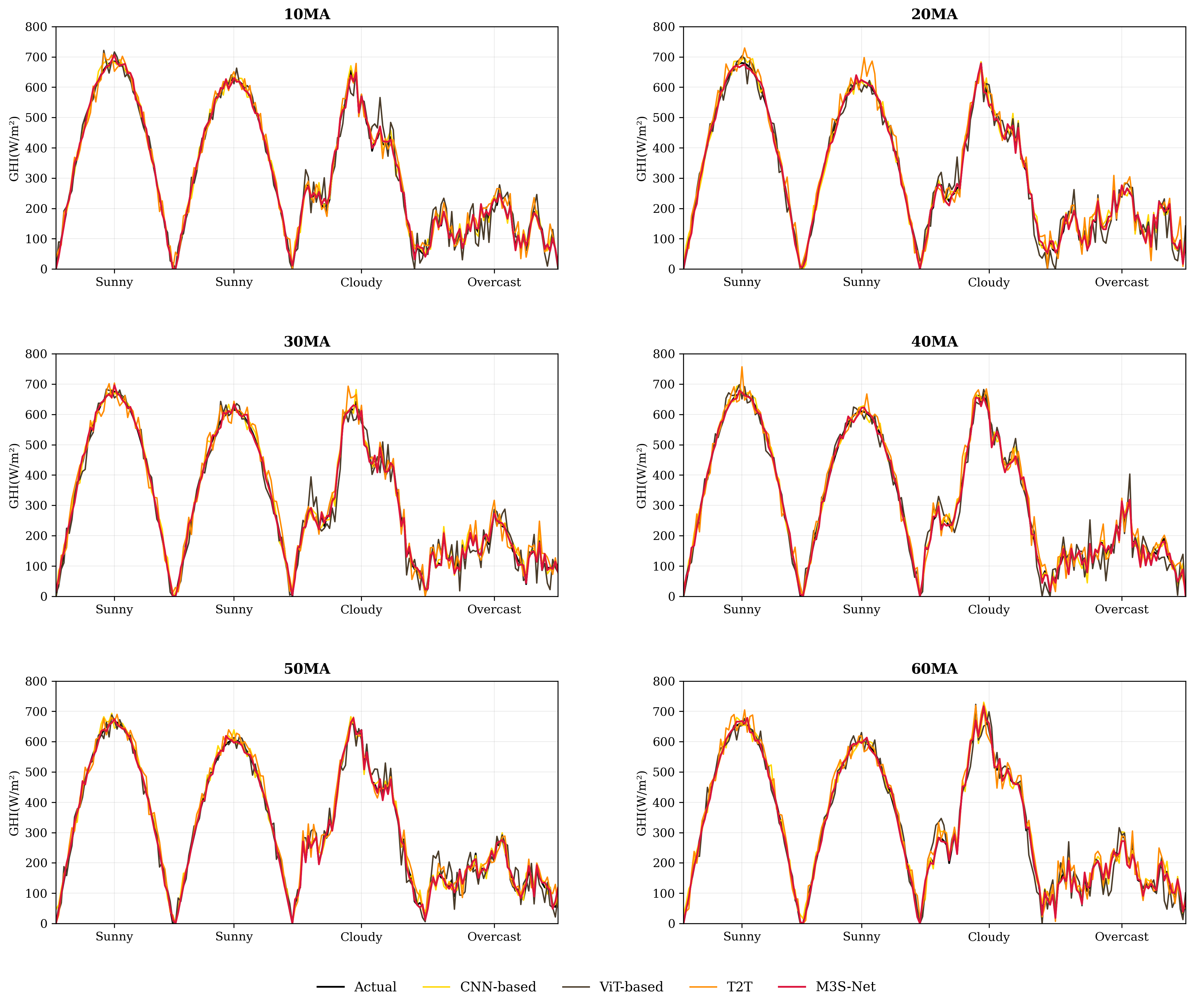}}
	\caption{ Visual results under three typical weather categories of the proposed M3S-Net and other multimodal methods.}
	\label{fig_11}
\end{figure*}

To investigate the model's generalization capabilities under diverse meteorological scenarios, the test dataset was stratified into three typical weather categories: sunny, cloudy, and overcast, as shown in Fig.~\ref{fig_10} and ~\ref{fig_11}. From the visual results, all models achieved satisfactory predictions under sunny conditions, where solar irradiance is stable and primarily governed by the deterministic solar trajectory. Consequently, temporal models alone suffice for reliable forecasting in these regimes. However, the M3S-Net prediction curve exhibited the tightest alignment with ground truth values and the lowest volatility. This substantiates the SIFR-Net module’s precision in extracting long-term stationary trends and diurnal periodic features. During ramping events characterized by rapid irradiance drops and recoveries due to cloud occlusion, multimodal time-series models, such as CNN-based and ViT-based, displayed significant forecasting error. Conversely, as the expansion of the prediction range, M3S-Net exhibits the fastest and most accurate response to steep power ramps, the error reduction of the fitting curve is the least significant compared to other models. This performance enhancement stems primarily from the MPCS-Net’s effective capture of fine-grained cloud boundaries and motion dynamics. Under overcast conditions, characterized by low overall irradiance and complex fluctuation patterns, M3S-Net maintained a relatively stable prediction performance. This indicates that its multi-scale feature extraction and fusion mechanisms can effectively adapt to varying irradiance intensities and weather modes.
\begin{table}[!t]
	\centering
	\caption{Ablation experimental results on the FGPD.}
	\resizebox{\linewidth}{!}{
		\begin{tabular}{cccc}
			\toprule
			Version & MAE ($\mathrm {W/m^2}$) ($\downarrow$) & NRMSE ($\%$) ($\downarrow$) & $\mathrm { R^{2}}$ ($\uparrow$)  \\
			\midrule
			A & 23.44 & 5.37 & 0.948 \\
			B & 22.46 & 5.21 & 0.951 \\
			C & 22.31 & 5.15 & 0.853 \\
			D & 21.84 & 5.06 & 0.957  \\
			E & 21.45 & 4.94 & 0.960  \\
			F & 20.96 & 4.72 & 0.962  \\
			M3S-Net & \textbf{19.84} & \textbf{4.61} & \textbf{0.964}  \\
			\bottomrule
	\end{tabular}}
	\label{Table6}
\end{table}
\subsection{Ablation Study}
To substantiate the necessity and individual contributions of the core components within M3S-Net, a rigorous series of ablation experiments was designed. All trials utilized an identical dataset comprising synchronized GCI and meteorological telemetry, maintaining consistent hyperparameter configurations to ensure the validity of the comparative analysis. The comparison results for the 10 MA are summarized in Table~\ref{Table6}. The experimental design segments the evaluation into six distinct schemes (A through F), each representing a specific architectural hypothesis.

Employing SIFR-Net to process historical power and meteorological data exclusively, we established a baseline (Scheme A) devoid of visual information. The MAE based on this configuration exceeded that of the complete model by approximately 15$\%$. This disparity implies that historical time-series data alone is insufficient for predicting rapid fluctuations induced by cloud movement.

Building upon Scheme A, the fine-grained visual extraction branch was substituted with a standard ResNet-50 (Scheme B) for feature extraction, fusing these visual features with time-series data via concatenation. This approach yielded performance enhancements over Scheme A, where the MAE decreased by approximately 4$\%$, thereby demonstrating the efficacy of incorporating visual data. However, its performance remained significantly inferior to the complete M3S-Net, suggesting that rudimentary feature extraction and fusion methods fail to fully exploit critical visual information or establish deep inter-modal interactions.

Substituting the cloud branch with our prior MPCM-Net architecture (Scheme C) resulted in performance superior to Scheme B but inferior to MPCS-Net equipped with the SCSM (Scheme D). These results indicate that the SCSM module, through its CSIA and CE mechanisms, adaptively enhances the representation of critical cloud features—such as thick cloud regions and motion fronts. Analysis of the MAE suggests this module contributes approximately 2$\%$ to the overall prediction accuracy.

In Scheme E, we utilized the complete MPCS-Net and SIFR-Net but eliminated the multi-scale ensemble prediction head within SIFR-Net, relying solely on the coarsest-scale features for the final forecast. Performance deteriorated, with a marked loss of accuracy in short-term forecasts. This finding substantiates the effectiveness of the multi-scale pyramid structure and ensemble prediction heads: fine-scale heads focus on capturing high-frequency, short-term fluctuations (e.g., gust effects), while coarse-scale heads model long-term trends and diurnal cycles. Their synergistic operation significantly enhances the model's adaptability to dynamics across varying temporal scales.

For Scheme F, we employed the full MPCS-Net and SIFR-Net but replaced the cross-modal Mamba fusion module with feature concatenation followed by a MLP. Experimental results indicate that this configuration performed significantly worse than the complete M3S-Net, with the $\mathrm {R^{2}}   $ for 10 MA forecasts decreasing by more than 0.02. This empirical evidence robustly substantiates the intrinsic value of the proposed cross-modal Mamba interaction mechanism. By exchanging state mapping matrices, this mechanism compels each modality to integrate context provided by the other during self-information decoding. This process facilitates genuine, sequence-wide deep feature complementarity and dependency modeling.

In conclusion, the proposed M3S-Net achieved superior performance across all evaluation protocols, thereby validating the rationality and advancement of the overall architectural design.

\subsection{Sensitivity Analysis}
To ensure model robustness and provide rigorous guidelines for practical deployment, we conducted a comprehensive sensitivity analysis of key hyperparameters governing the M3S-Net architecture.

\begin{figure}[!t]
	\centering{\includegraphics[width=3in]{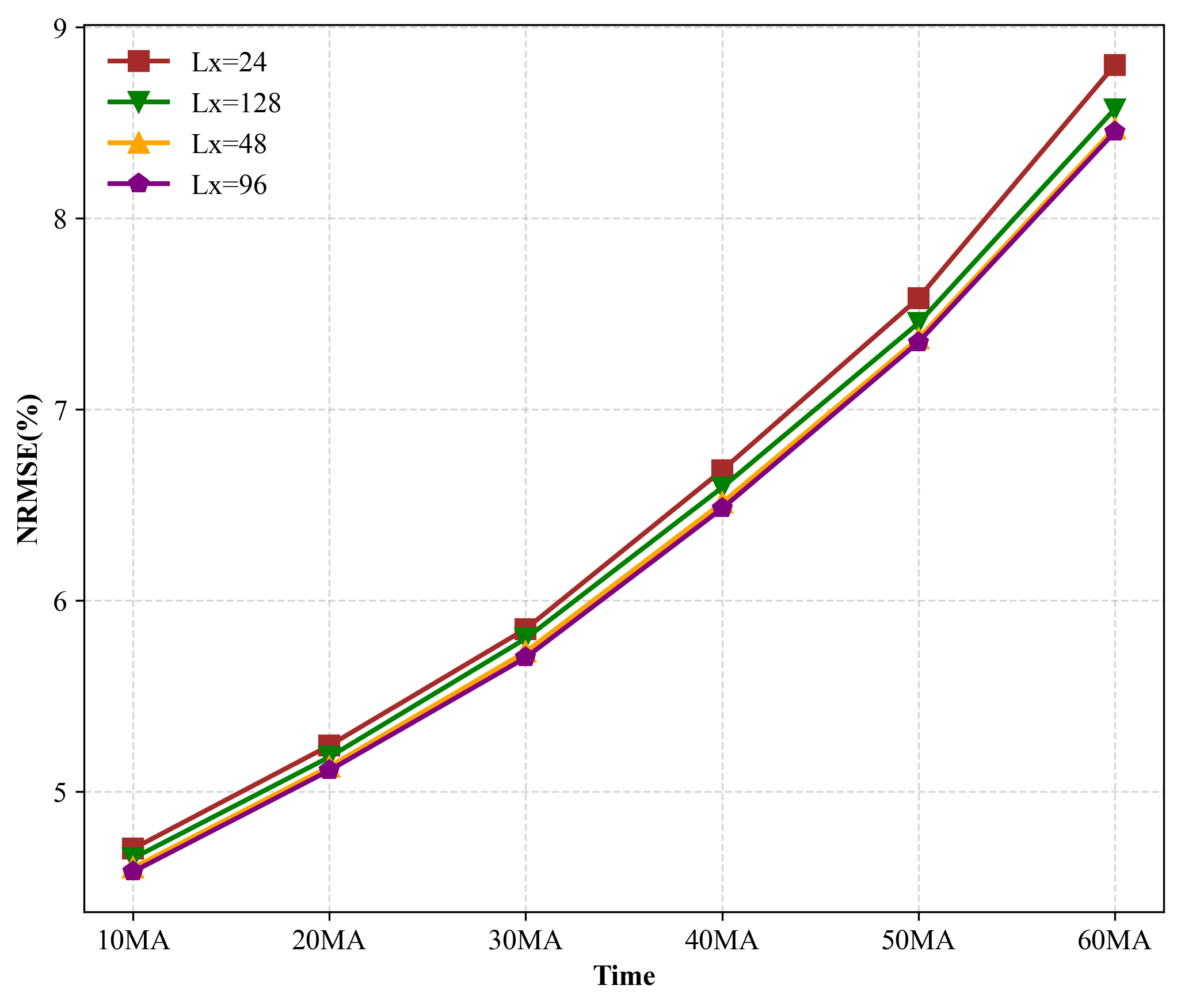}}
	\caption{ Visualization of the impact of input lengths in SIFR-Net.}
	\label{fig_12}
\end{figure}

First, we investigated the influence of the historical input sequence length, denoted as the lookback window $L_x$. We evaluated the model's performance with input lengths of 24, 48, 96, and 192 time steps, corresponding to historical horizons of 4, 8, 16, and 32 hours, respectively. As illustrated in Fig.~\ref{fig_12}, the model achieves optimal predictive accuracy across the majority of forecast horizons when $L_x = 96$ (16 hours).

\begin{figure}[!t]
	\centering{\includegraphics[width=3in]{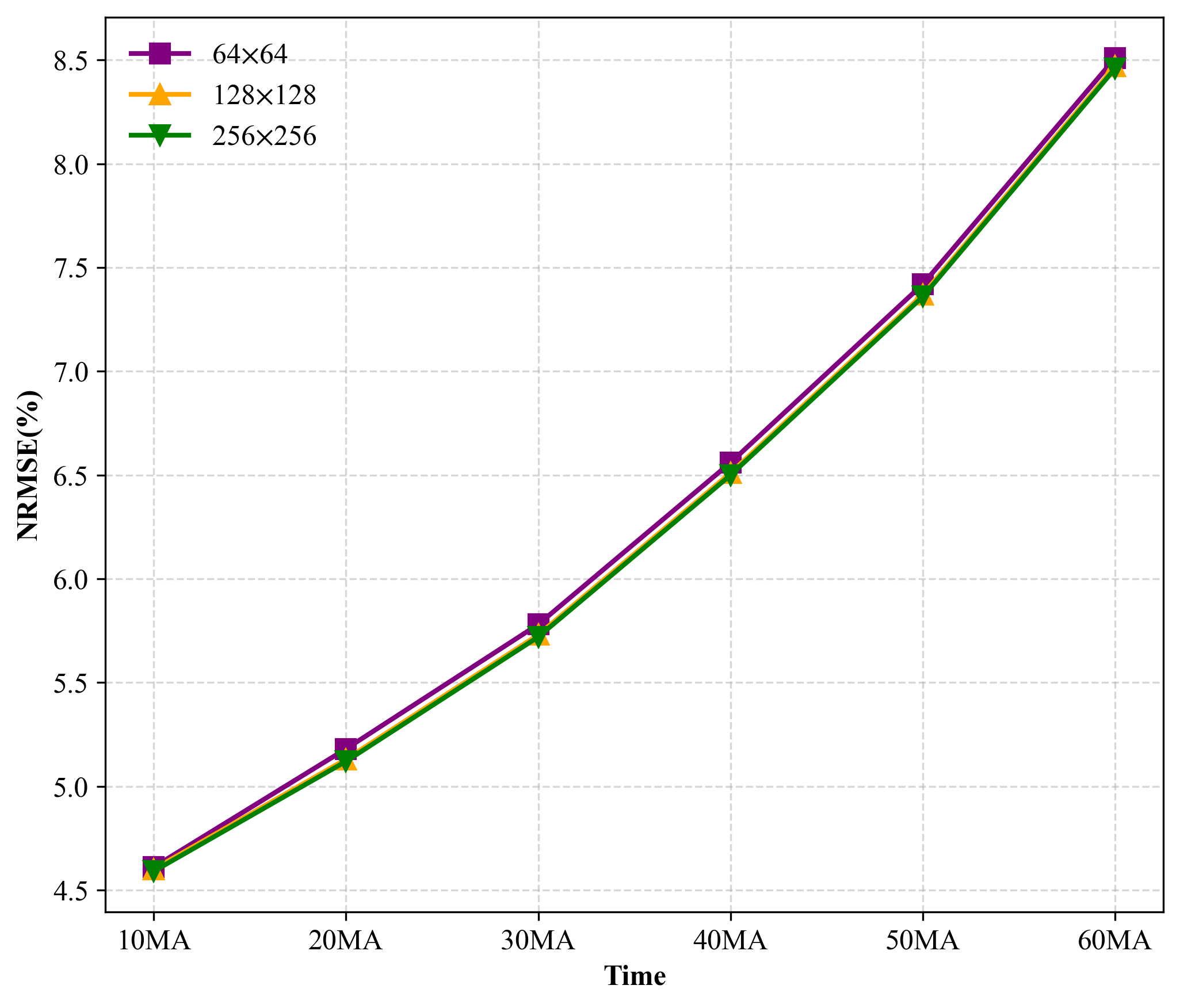}}
	\caption{ Visualization of the impact of image resolution in MPCS-Net.}
	\label{fig_13}
\end{figure}

Subsequently, we examined the effect of the input resolution of ground-based sky images, testing configurations of $64 \times 64$, $128 \times 128$, and $256 \times 256$ pixels. The results, presented in Fig.~\ref{fig_13}, indicate that the impact of resolution is minimal on our M3S-Net. Consequently, we use image resolution $64 \times 64$ for GHI forecasting in this paper.
\begin{table}[!t]
	\centering
	\caption{Comparison results of GHI with different methods on the SRRL dataset ($\%$).}
	\resizebox{\linewidth}{!}{
		\begin{tabular}{cccccccc}
			\toprule
			\multirow{2}{*}{\makecell[c]{Model \\structure}} & \multirow{2}{*}{Method} & \multicolumn{6}{c}{NRMSE ($\%$)} \\ 
			\cmidrule(lr){3-8}
			& & \makecell[c]{10 \\MA} & \makecell[c]{20 \\MA} & \makecell[c]{30 \\MA} & \makecell[c]{40 \\MA} & \makecell[c]{50 \\MA} & \makecell[c]{60 \\MA} \\
			\midrule
			
			\multirow{3}{*}{\makecell[c]{Single\\-modal}} 
			& LSTM     & 7.73 & 7.83 & 8.12 & 8.64 & 9.37 & 10.01 \\
			& DLinear  & 4.72 & 6.88 & 8.38 & 9.62 & 10.84 & 12.03 \\
			& TimesNet & 5.12 & 6.39 & 7.29 & 8.15 & 8.99 & 9.84 \\
			
			\multirow{4}{*}{\makecell[c]{Multi\\-modal}} 
			& CNN-based   & 5.03 & 6.19 & 7.08 & 7.88 & 8.71 & 9.33 \\
			& ViT-based      & 5.04 & 6.16 & 7.02 & 7.92 & 8.66 & 9.36 \\
			& T2T & 4.64 & 5.98 & 6.88 & 7.68 & 8.24 & 8.83 \\
			& M3S-Net & \textbf{4.35} & \textbf{5.79} & \textbf{6.76} & \textbf{7.61} & \textbf{8.16} & \textbf{8.72} \\
			\bottomrule
	\end{tabular}}
	\label{Table7}
\end{table}

\subsection{Generalization Ability Analysis}
To rigorously position M3S-Net within the broader research landscape, we benchmarked its predictive performance against representative recent studies utilizing comparable public datasets (e.g., NREL SRRL) and forecast horizons. To ensure equitable comparison, the NRMSE was adopted as the unified evaluation metric. Table~\ref{Table7} summarizes the methodologies, forecast horizons, and reported NRMSE values for these related works. Ref. \cite{CNN-based} employed a CNN backbone for feature extraction of series, reporting a 10-minute forecast NRMSE of 6.14$\%$. While effective for spatial features, this approach lacks explicit temporal modeling of cloud dynamics. 
\begin{figure}[!t]
	\centering{\includegraphics[width=3in]{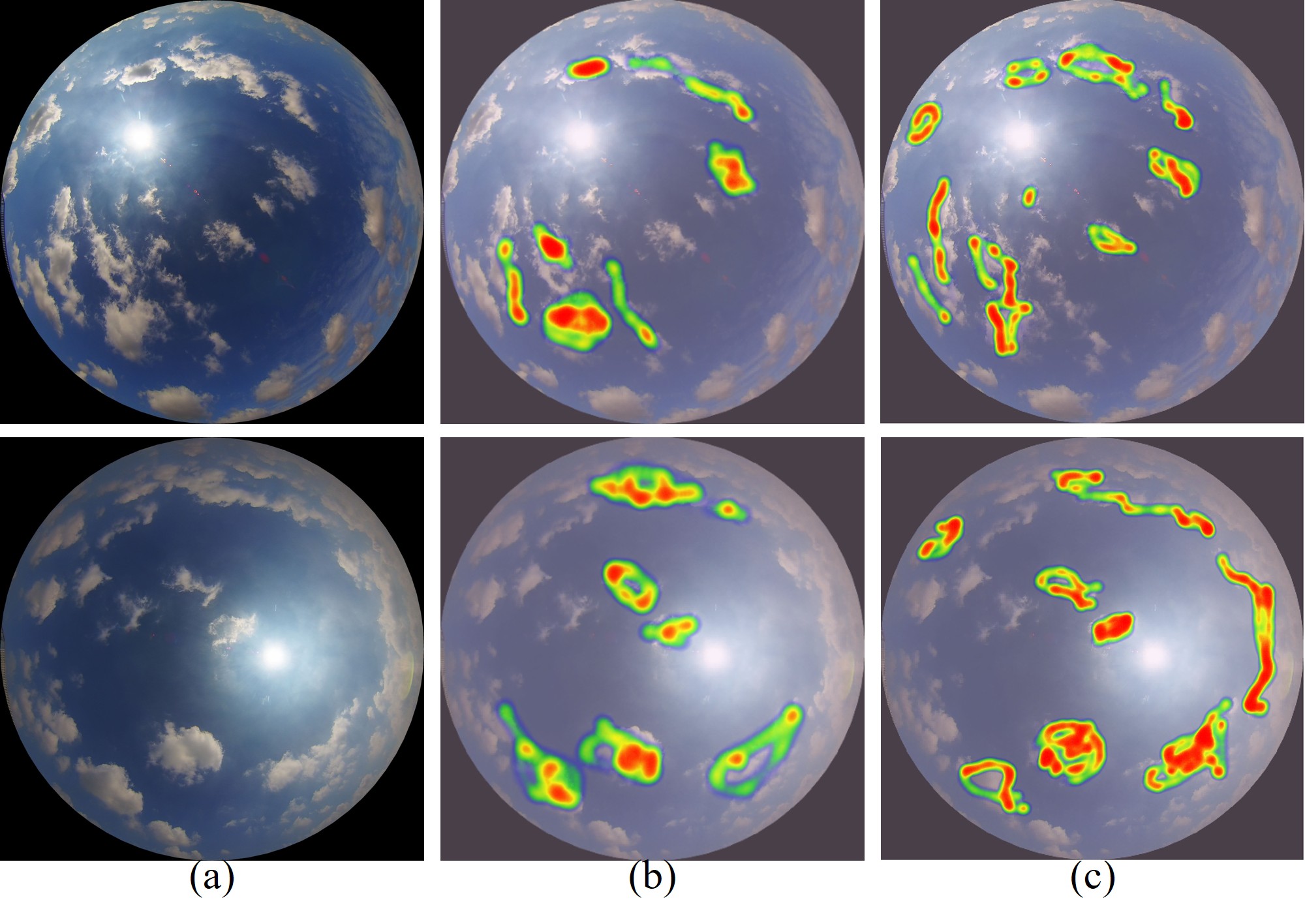}}
	\caption{ Visualization results show different categories of the FGPD validation set using Grad-CAM. (a) Input image, (b) MPCM-Net, and (c) MPCS-Net.}
	\label{fig_14}
\end{figure}

Li et al. \cite{T2T-ViT} converted the historical sequence into a 2D tensor using TimesNet and extracted temporal features using CNN, achieving NRMSE values between 4.64$\%$ and 6.88$\%$ for 10-to-30-minute horizons. In contrast, our proposed M3S-Net achieves a significantly lower NRMSE of 4.64$\%$ for 10-minute forecasts on comparable datasets. Furthermore, M3S-Net demonstrates superior or highly competitive performance across the entire 10-to-60-minute forecast range. This performance advantage is attributed to the synergistic integration of fine-grained cloud segmentation (MPCS-Net), comprehensive multi-scale temporal modeling (SIFR-Net), and the deep cross-modal fusion mechanism (MMIF) inherent to our architecture.

\begin{figure*}[!t]
	\centering{\includegraphics[width=6in]{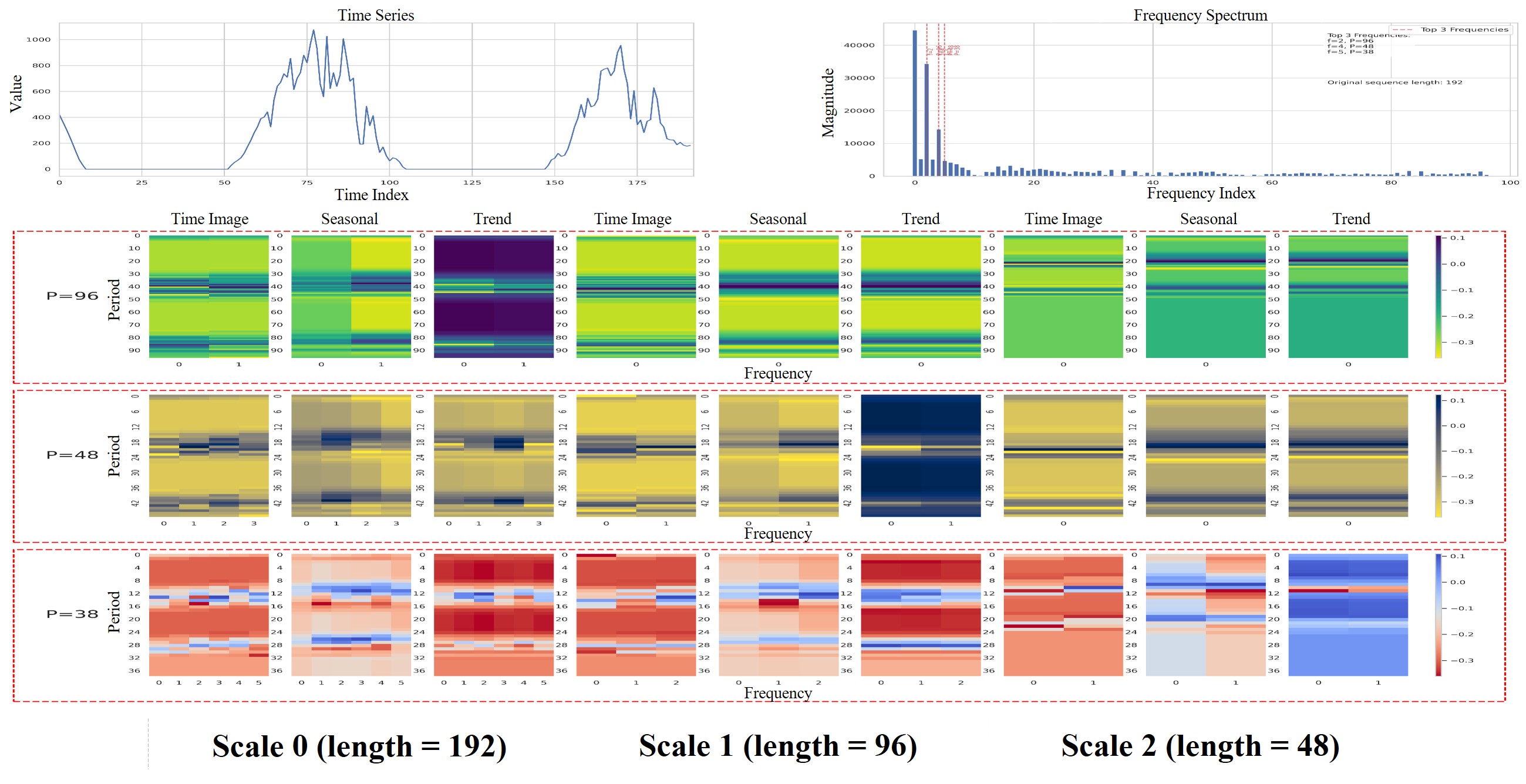}}
	\caption{Visualization of representation on FGPD across three different scales (length: 192, 96, 48) and three periods (96, 48, 36) with frequencies (2, 4, 5). Both high-frequency local fluctuations and low-frequency seasonal trends via spectral-temporal disentanglement are captured.}
	\label{fig_15}
\end{figure*}
\begin{figure*}[!t]
	\centering{\includegraphics[width=6in]{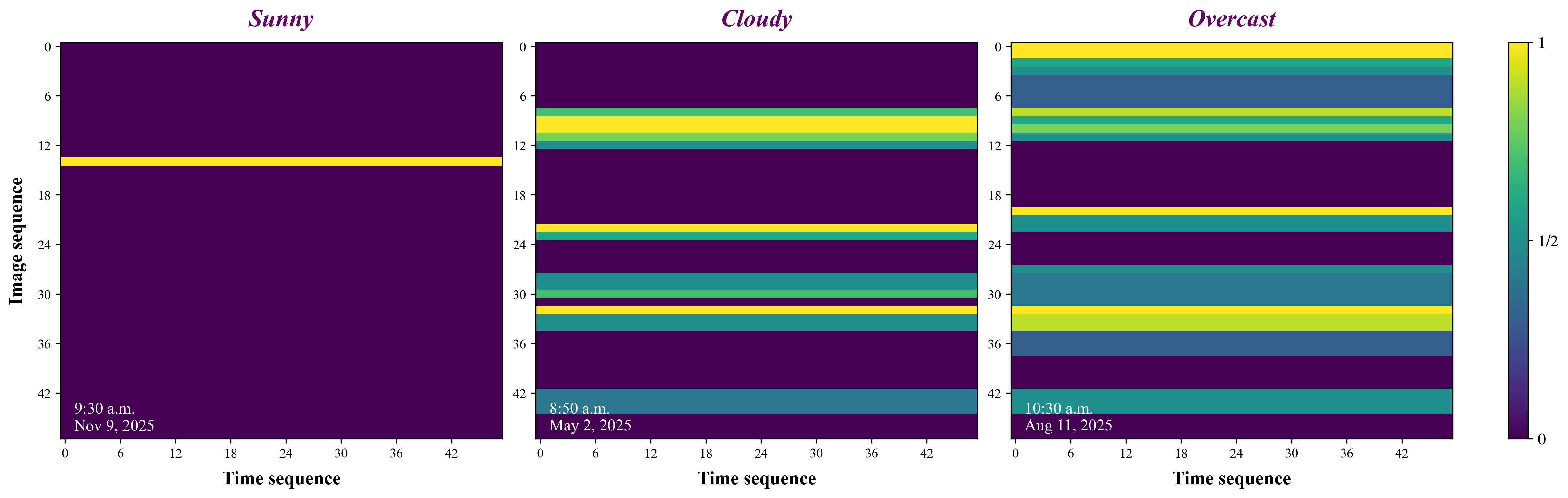}}
	\caption{The heatmaps of visual and temporal weights of MMIF under different weather conditions. Highlight performance along the horizontal axis indicates the difference in the visual feature activation. Visual feature is activated to enhance temporal features to participate in GHI forecasting.}
	\label{fig_16}
\end{figure*}
\subsection{Model Interpretability and Visualization}
To enhance model transparency and verify the physical meaningfulness of the learned features, we conducted a comprehensive visualization analysis of M3S-Net’s key components. We utilized Gradient-weighted Class Activation Mapping (Grad-CAM) to generate heatmaps for the MPCS-Net module when processing typical multi-cloud images, as shown in Fig.~\ref{fig_14}. The heatmaps reveal that during fine-grained visual extraction branch, the model focuses its attention intensely on cloud regions actively moving toward or occluding the sun, specifically highlighting cloud edges and optically thick areas. This intuitive verification confirms that MPCS-Net successfully learns spatial features directly correlated with irradiance attenuation, rather than relying on spurious background correlations. By applying FFT to the input power sequence, we extracted dominant periodic components. Fig.~\ref{fig_15} demonstrates that SIFR-Net successfully reshapes 1D time-series data into 2D tensor structures based on these dominant periods (e.g., diurnal cycles, short-term fluctuations). The module's scalable attention mechanism then learns intra-period variations and inter-period evolution along the row and column dimensions, respectively, yielding physically meaningful feature representations. Finally, we extracted the interaction attention weights from the cross-modal Mamba layer in the MMIF module. Fig.~\ref{fig_16} displays the weight heatmaps for visual and temporal features under sunny, cloudy, and overcast conditions. Under sunny conditions, the weights assigned to visual features are minimal, indicating that the model relies on the temporal regularity of the clear-sky curve. In contrast, under cloudy conditions, the visual weights become significantly more prominent, demonstrating the model's adaptability to visual data for predicting fluctuation trends. Weights in overcast remain consistently volatile, suggesting a continuous reliance on visual data to interpret complex, high-entropy sky conditions. These qualitative results confirm that the fusion module functions as a dynamic gate, modulating modal importance in response to real-time environmental contexts.

\section{Conclusion}\label{sec:conclusion}
This study advances the domain of ultra-short-term PV power forecasting by transcending the inherent limitations of conventional binary cloud segmentation and shallow feature concatenation techniques. The proposed M3S-Net framework establishes a robust cross-modal semantic alignment between ground-based cloud imagery and meteorological time-series data. Specifically, by deploying the MPCS-Net to capture fine-grained cloud optical depth and the SIFR-Net to disentangle temporal periodicity, the framework effectively bridges the heterogeneous modality gap. Central to this innovation is the MMIF module, which incorporates a dynamic state-space matrix switching mechanism. This mechanism enables temporal contexts to adaptively modulate the representation of visual features. Extensive empirical evaluations conducted on the newly FGPD and NREL benchmark datasets substantiate the framework's performance, demonstrating significant improvements over existing SOTA baselines. However, acknowledging the framework's reliance on high-quality, fine-grained annotations, future investigations will prioritize semi-supervised or self-supervised domain adaptation techniques to enhance label efficiency and model generalization.

\section*{CRediT authorship contribution statement}\label{sec: CR}
\textbf{Penghui Niu:} Writing - original draft, Writing – review $\&$ editing, Visualization, Validation, Software, Methodology, Investigation, Funding acquisition. \textbf{Taotao Cai:} Writing – review $\&$ editing, Methodology, Investigation, Formal analysis. 
\textbf{Suqi Zhang:} Supervision, Resources, Project administration, Data curation. 
\textbf{Junhua Gu:} Supervision, Resources, Project administration, Funding acquisition, Data curation. \textbf{Ping Zhang:} Resources, Project administration, Funding acquisition.
\textbf{Qiqi Liu:} Resources, Project administration, Investigation. 
\textbf{Jianxin Li:} Writing – review $\&$ editing, Supervision, Project administration, Methodology.

\section*{Declaration of competing interest}\label{sec: Dec}
The authors declare that they have no known competing finan cial interests or personal relationships that could have appeared to influence the work reported in this paper.

\section*{Data availability}\label{sec: data}
The proposed FGPD dataset and source code will be available at https://github.com/she1110/FGPD.

\section*{Acknowledgments}\label{sec: ack}
The research was supported by the National Natural Science Foundation of China under Grant No. 62206085, supported by the Innovation Capacity Improvement Plan Project of Hebei Province under 22567603H, and supported by the Interdisciplinary Postgraduate Training Program of Hebei University of Technology under HEBUT-Y-XKJC-2022101.

% Use the style of numbering in square brackets.
% If nothing is used, default style will be taken.
%\begin{enumerate}[a)]
%\item 
%\item 
%\item 
%\end{enumerate}  

%\begin{itemize}
%\item 
%\item 
%\item 
%\end{itemize}  

%\begin{description}
%\item[]
%\item[] 
%\item[] 
%\end{description}  

% Figure
% \begin{figure}[<options>]
% 	\centering
% 		\includegraphics[<options>]{}
% 	  \caption{}\label{fig1}
% \end{figure}

% \begin{table}[<options>]
% \begin{tabular*}{\tblwidth}{@{}LL@{}}
% \toprule
% \midrule
%  & \\
%  & \\
%  & \\
%  & \\
% \bottomrule
% \end{tabular*}
% \end{table}

% Uncomment and use as the case may be
%\begin{theorem} 
%\end{theorem}

% Uncomment and use as the case may be
%\begin{lemma} 
%\end{lemma}



%\bibliographystyle{cas-model2-names}
%\bibliographystyle{elsarticle-harv}
\bibliographystyle{elsarticle-num}

\bibliography{reference}

@article{Intro1,
title = {Multi-step prediction of photovoltaic power based on two-stage decomposition and BILSTM},
journal = {Neurocomputing},
volume = {504},
pages = {56-67},
year = {2022},
issn = {0925-2312},
doi = {https://doi.org/10.1016/j.neucom.2022.06.117},
author = {Wenshuai Lin and Bin Zhang and Hongyi Li and Renquan Lu}}

@article{Intro2,
title = {Solar energy: Potential and future prospects},
journal = {Renewable Sustainable Energy Rev.},
volume = {82},
pages = {894-900},
year = {2018},
issn = {1364-0321},
doi = {https://doi.org/10.1016/j.rser.2017.09.094},
author = {Ehsanul Kabir and Pawan Kumar and Sandeep Kumar and Adedeji A. Adelodun and Ki-Hyun Kim}
}

@article{Intro3,
title = {Using sky-classification to improve the short-term prediction of irradiance with sky images and convolutional neural networks},
journal = {Sol. Energy},
volume = {269},
pages = {112320},
year = {2024},
issn = {0038-092X},
doi = {https://doi.org/10.1016/j.solener.2024.112320},
author = {Victor Arturo {Martinez Lopez} and Gijs {van Urk} and Pim J.F. Doodkorte and Miro Zeman and Olindo Isabella and Hesan Ziar}
}

@article{Intro4,
title = {An evolutionary robust solar radiation prediction model based on WT-CEEMDAN and IASO-optimized outlier robust extreme learning machine},
journal = {Appl. Energy},
volume = {322},
pages = {119518},
year = {2022},
issn = {0306-2619},
doi = {https://doi.org/10.1016/j.apenergy.2022.119518},
author = {Chu Zhang and Lei Hua and Chunlei Ji and Muhammad {Shahzad Nazir} and Tian Peng}
}

@article{Intro5,
title = {Challenges and solution technologies for the integration of variable renewable energy sources—a review},
journal = {Renewable Energy},
volume = {145},
pages = {2271-2285},
year = {2020},
issn = {0960-1481},
doi = {https://doi.org/10.1016/j.renene.2019.06.147},
author = {Simon R. Sinsel and Rhea L. Riemke and Volker H. Hoffmann}
}

@article{Intro6,
title = {Numerical weather prediction (NWP) and hybrid ARMA/ANN model to predict global radiation},
journal = {Energy},
volume = {39},
number = {1},
pages = {341-355},
year = {2012},
note = {Sustainable Energy and Environmental Protection 2010},
issn = {0360-5442},
doi = {https://doi.org/10.1016/j.energy.2012.01.006},
author = {Cyril Voyant and Marc Muselli and Christophe Paoli and Marie-Laure Nivet}
}

@article{Intro7,
title = {Hourly forecasting of solar irradiance based on CEEMDAN and multi-strategy CNN-LSTM neural networks},
journal = {Renewable Energy},
volume = {162},
pages = {1665-1683},
year = {2020},
issn = {0960-1481},
doi = {https://doi.org/10.1016/j.renene.2020.09.141},
author = {Bixuan Gao and Xiaoqiao Huang and Junsheng Shi and Yonghang Tai and Jun Zhang}
}

@article{Intro8,
  author       = {Chao Huang and
                  Long Wang and
                  Loi Lei Lai},
  title        = {Data-Driven Short-Term Solar Irradiance Forecasting Based on Information
                  of Neighboring Sites},
  journal      = {{IEEE} Trans. Ind. Electron.},
  volume       = {66},
  number       = {12},
  pages        = {9918--9927},
  year         = {2019},
  doi          = {10.1109/TIE.2018.2856199},
  bibsource    = {dblp computer science bibliography, https://dblp.org}
}

@article{Intro9,
Author = {Xia, Pan and Zhang, Lu and Min, Min and Li, Jun and Wang, Yun and Yu, Yu
   and Jia, Shengjie},
Title = {Accurate nowcasting of cloud cover at solar photovoltaic plants using
   geostationary satellite images},
Journal = {Nat. Commun.},
Year = {2024},
Volume = {15},
Number = {1},
Month = {JAN 13},
DOI = {10.1038/s41467-023-44666-1}
}

@article{Intro10,
title = {A ground-based cloud image classification method for photovoltaic power prediction based on Convolutional Neural Networks and Vision Transformer},
journal = {Eng. Appl. Artif. Intell.},
volume = {159},
pages = {111582},
year = {2025},
issn = {0952-1976},
doi = {https://doi.org/10.1016/j.engappai.2025.111582},
author = {Chaojun Shi and Mengyu Zhang and Hongyin Xiang and Ke Zhang and Sihao Ju and Xiaoyun Zhang and Leile Han}
}

@article{Intro11,
title = {A hybrid ensemble optimized BiGRU method for short-term photovoltaic generation forecasting},
journal = {Energy},
volume = {299},
pages = {131458},
year = {2024},
issn = {0360-5442},
doi = {https://doi.org/10.1016/j.energy.2024.131458},
author = {Yeming Dai and Weijie Yu and Mingming Leng}
}

@article{Intro12,
title = {Hybrid ultra-short-term PV power forecasting system for deterministic forecasting and uncertainty analysis},
journal = {Energy},
volume = {288},
pages = {129898},
year = {2024},
issn = {0360-5442},
doi = {https://doi.org/10.1016/j.energy.2023.129898},
author = {Jianzhou Wang and Yue Yu and Bo Zeng and Haiyan Lu}
}

@article{Intro13,
title = {Multistage spatio-temporal attention network based on NODE for short-term PV power forecasting},
journal = {Energy},
volume = {290},
pages = {130308},
year = {2024},
issn = {0360-5442},
doi = {https://doi.org/10.1016/j.energy.2024.130308},
author = {Songtao Huang and Qingguo Zhou and Jun Shen and Heng Zhou and Binbin Yong}
}

@article{Intro14,
title = {Photovoltaic power electricity generation nowcasting combining sky camera images and learning supervised algorithms in the Southern Spain},
journal = {Renewable Energy},
volume = {206},
pages = {251-262},
year = {2023},
issn = {0960-1481},
doi = {https://doi.org/10.1016/j.renene.2023.01.111},
author = {Mauricio Trigo-González and Marcelo Cortés-Carmona and Aitor Marzo and Joaquín Alonso-Montesinos and Mercedes Martínez-Durbán and Gabriel López and Carlos Portillo and Francisco Javier Batlles}
}

@article{Intro15,
title = {Convolutional neural networks for intra-hour solar forecasting based on sky image sequences},
journal = {Appl. Energy},
volume = {310},
pages = {118438},
year = {2022},
issn = {0306-2619},
doi = {https://doi.org/10.1016/j.apenergy.2021.118438},
author = {Cong Feng and Jie Zhang and Wenqi Zhang and Bri-Mathias Hodge}
}

@article{Intro16,
title = {Short-term solar radiation forecasting with a novel image processing-based deep learning approach},
journal = {Renewable Energy},
volume = {200},
pages = {1490-1505},
year = {2022},
issn = {0960-1481},
doi = {https://doi.org/10.1016/j.renene.2022.10.063},
author = {Ardan Hüseyin Eşlik and Emre Akarslan and Fatih Onur Hocaoğlu}}

@article{Intro17,
Author = {Liu, Jingxuan and Zang, Haixiang and Ding, Tao and Cheng, Lilin and Wei,
   Zhinong and Sun, Guoqiang},
Title = {Sky-Image-Derived Deep Decomposition for Ultra-Short-Term Photovoltaic
   Power Forecasting},
Journal = {{IEEE} Trans. Sustainable Energy},
Year = {2024},
Volume = {15},
Number = {2},
Pages = {871-883},
DOI = {10.1109/TSTE.2023.3312401}
}

@article{Intro18,
title = {Improving ultra-short-term photovoltaic power forecasting using a novel sky-image-based framework considering spatial-temporal feature interaction},
journal = {Energy},
volume = {293},
pages = {130538},
year = {2024},
issn = {0360-5442},
doi = {https://doi.org/10.1016/j.energy.2024.130538},
author = {Haixiang Zang and Dianhao Chen and Jingxuan Liu and Lilin Cheng and Guoqiang Sun and Zhinong Wei}
}

@article{Kong,
title = {Hybrid approaches based on deep whole-sky-image learning to photovoltaic generation forecasting},
journal = {Appl. Energy},
volume = {280},
pages = {115875},
year = {2020},
issn = {0306-2619},
doi = {https://doi.org/10.1016/j.apenergy.2020.115875},
author = {Weicong Kong and Youwei Jia and Zhao Yang Dong and Ke Meng and Songjian Chai}
}

@ARTICLE{Dolatabadi,
  author={Dolatabadi, Amirhossein and Abdeltawab, Hussein and Mohamed, Yasser Abdel-Rady I.},
  journal={{IEEE} Trans. Power Syst.}, 
  title={Deep Reinforcement Learning-Based Self-Scheduling Strategy for a CAES-PV System Using Accurate Sky Images-Based Forecasting}, 
  year={2023},
  volume={38},
  number={2},
  pages={1608-1618},
  doi={10.1109/TPWRS.2022.3177704}}

@article{Terr,
title = {Deep learning for intra-hour solar forecasting with fusion of features extracted from infrared sky images},
journal = {Inf. Fusion},
volume = {95},
pages = {42-61},
year = {2023},
issn = {1566-2535},
doi = {https://doi.org/10.1016/j.inffus.2023.02.006},
author = {Guillermo Terrén-Serrano and Manel Martínez-Ramón}
}

@article{Caldas,
title = {Very short-term solar irradiance forecast using all-sky imaging and real-time irradiance measurements},
journal = {Renewable Energy},
volume = {143},
pages = {1643-1658},
year = {2019},
issn = {0960-1481},
doi = {https://doi.org/10.1016/j.renene.2019.05.069},
author = {M. Caldas and R. Alonso-Suárez}
}

@article{Liu,
title = {Harvesting spatiotemporal correlation from sky image sequence to improve ultra-short-term solar irradiance forecasting},
journal = {Renewable Energy},
volume = {209},
pages = {619-631},
year = {2023},
issn = {0960-1481},
doi = {https://doi.org/10.1016/j.renene.2023.03.122},
author = {Jingxuan Liu and Haixiang Zang and Tao Ding and Lilin Cheng and Zhinong Wei and Guoqiang Sun}
}

@ARTICLE{Wu,
  author={Wu, Xinyue and Zhen, Zhao and Zhang, Jianmei and Wang, Fei and Xu, Fei and Ren, Hui and Su, Ying and Sun, Yong and Yang, Heng},
  journal={{IEEE} Trans. Ind. Appl.}, 
  title={Multidimensional Feature Extraction Based Minutely Solar Irradiance Forecasting Method Using All-Sky Images}, 
  year={2024},
  volume={60},
  number={3},
  pages={4494-4504},
  doi={10.1109/TIA.2024.3372515}}

@article{T2T-ViT,
title = {Multi-modal feature fusion model based on TimesNet and T2T-ViT for ultra-short-term solar irradiance forecasting},
journal = {Renewable Energy},
volume = {240},
pages = {122192},
year = {2025},
issn = {0960-1481},
doi = {https://doi.org/10.1016/j.renene.2024.122192},
author = {Hao Li and Gang Ma and Bo Wang and Shu Wang and Wenhao Li and Yuxiang Meng}
}

@article{Dou,
title = {A multi-modal deep clustering method for day-ahead solar irradiance forecasting using ground-based cloud imagery and time series data},
journal = {Energy},
volume = {321},
pages = {135285},
year = {2025},
issn = {0360-5442},
doi = {https://doi.org/10.1016/j.energy.2025.135285},
author = {Weijing Dou and Kai Wang and Shuo Shan and Mingyu Chen and Kanjian Zhang and Haikun Wei and Victor Sreeram}
}

@article{Zhu,
title = {Hybrid modeling of direct normal irradiance for rooftop photovoltaic systems using multi-feature extraction from cloud images},
journal = {J. Build. Eng.},
volume = {106},
pages = {112571},
year = {2025},
issn = {2352-7102},
doi = {https://doi.org/10.1016/j.jobe.2025.112571},
author = {Junye Zhu and Yangshu Lin and Chao Yang and Keqi Wang and Zhongwei Zhang and Zhongyang Zhao and Lijie Wang and Zhiming Lin and Qiwen Jin and Chenghang Zheng and Xuecheng Wu and Xiang Gao}
}

@article{Wei,
  author       = {Xiangsen Wei and
                  Dong Yue and
                  Gerhard P. Hancke and
                  Chunxia Dou and
                  Houjun Li and
                  Yang Qiu},
  title        = {Ultra Short-Term Solar Irradiance Forecast Based on Multimodal Data
                  Fusion and Fuzzification},
  journal      = {{IEEE} Trans. Ind. Informatics},
  volume       = {21},
  number       = {4},
  pages        = {3256--3265},
  year         = {2025},
  doi          = {10.1109/TII.2024.3523575}
}

@article{Nie,
  author       = {Bing Nie and
                  Zhiying Lu and
                  Jun Han and
                  Wenpeng Chen and
                  Chao Cai and
                  Wenjie Pan},
  title        = {Investigation on Ground-Based Cloud Image Classification and Its Application
                  in Photovoltaic Power Forecasting},
  journal      = {{IEEE} Trans. Instrum. Meas.},
  volume       = {74},
  pages        = {1--11},
  year         = {2025},
  doi          = {10.1109/TIM.2025.3529074}
}

@article{Ma,
title = {Research on ultra-short-term photovoltaic power forecasting using multimodal data and ensemble learning},
journal = {Energy},
volume = {330},
pages = {136831},
year = {2025},
issn = {0360-5442},
doi = {https://doi.org/10.1016/j.energy.2025.136831},
author = {Yifeng Ma and Wenzheng Yu and Junyu Zhu and Zhiyuan You and Aiqing Jia}
}

@article{Wei2,
title = {Climate-informed long-term forecasting of wind and photovoltaic power using a hybrid DWT–BES–CNN–LSTM model},
journal = {Energy},
volume = {338},
pages = {138677},
year = {2025},
issn = {0360-5442},
doi = {https://doi.org/10.1016/j.energy.2025.138677},
author = {Xingchen Wei and Xinyu Wu and Kei Yoshimura and Chuntian Cheng and Hao Huang and Zhendong Ding and Yuhang Song}
}

@article{Xiang,
title = {A dynamic meteorological correlation integrated hybrid method for photovoltaic output forecasting},
journal = {Renewable Energy},
volume = {256},
pages = {124561},
year = {2026},
issn = {0960-1481},
doi = {https://doi.org/10.1016/j.renene.2025.124561},
author = {Yue Xiang and Yunjie Yang and Lixiong Xu and Zhiyuan Tang and Youbo Liu and Wei Sun and Junyong Liu}
}

@article{Shi,
title = {A ground-based cloud image classification method for photovoltaic power prediction based on Convolutional Neural Networks and Vision Transformer},
journal = {Eng. Appl. Artif. Intell.},
volume = {159},
pages = {111582},
year = {2025},
issn = {0952-1976},
doi = {https://doi.org/10.1016/j.engappai.2025.111582},
author = {Chaojun Shi and Mengyu Zhang and Hongyin Xiang and Ke Zhang and Sihao Ju and Xiaoyun Zhang and Leile Han}
}

@article{ConvODE-Mixer,
title = {ConvODE-Mixer: A multimodal deep learning model for ultra-short-term PV power forecasting},
journal = {Sol. Energy},
volume = {300},
pages = {113777},
year = {2025},
issn = {0038-092X},
doi = {https://doi.org/10.1016/j.solener.2025.113777},
author = {Binbin Yong and Yanxiang Zhang and Jun Shen and Aiai Ren and Xu Zhou and Qingguo Zhou}
}

@article{Shi2,
title = {TransFCloudNet: a dual-branch feature fusion ground-based cloud image fine-grained segmentation method for photovoltaic power prediction},
journal = {Energy Convers. Manage.},
volume = {348},
pages = {120647},
year = {2026},
issn = {0196-8904},
doi = {https://doi.org/10.1016/j.enconman.2025.120647},
author = {Zibo Su and Chaojun Shi and Ke Zhang and Xiongbin Xie and Xiaoyun Zhang and Junchi Xiao}
}

@techreport{SRRL,
  title={Nrel solar radiation research laboratory (srrl): Baseline measurement system (bms); golden, colorado (data)},
  author={Stoffel, T and Andreas, A},
  year={1981},
  institution={National Renewable Energy Lab.(NREL), Golden, CO (United States)},
  DOI={http://dx.doi.org/10.7799/1052221,}
}

@article{Folsom,
Author = {Pedro, Hugo T. C. and Larson, David P. and Coimbra, Carlos F. M.},
Title = {A comprehensive dataset for the accelerated development and benchmarking
   of solar forecasting methods},
Journal = {J. Renewable Sustainable Energy},
Year = {2019},
Volume = {11},
Number = {3},
pages = {036102 },
DOI = {10.1063/1.5094494}
}

@article{SIRTA,
Author = {Haeffelin, M. and Barthès, L. and Bock, O. and Boitel, C. and Bony, S. and
 Bouniol, D. and Chepfer, H. and Chiriaco, M. and Cuesta, J. and Delanoë, J. and
 Drobinski, P. and Dufresne, J.-L. and
 Flamant, C. and Grall, M. and Hodzic, A. and Hourdin, F. and
 Lapouge, F. and Lemaître, Y. and Mathieu, A. and Morille, Y. and Naud, C. and
 Noël, V. and O'Hirok, W. and Pelon, J. and Pietras, C. and Protat, A. and Romand, B. and Scialom, G. and Vautard, R.},
Title = {SIRTA, a ground-based atmospheric observatory for cloud and aerosol research},
Journal = {Ann. Geophys.},
Year = {2005},
Volume = {23},
Number = {2},
pages = {253-275},
DOI = {https://doi.org/10.5194/angeo-23-253-2005}
}

@Article{WSISEG,
AUTHOR = {Xie, W. and Liu, D. and Yang, M. and Chen, S. and Wang, B. and Wang, Z. and Xia, Y. and Liu, Y. and Wang, Y. and Zhang, C.},
TITLE = {SegCloud: a novel cloud image segmentation model using a deep convolutional
neural network for ground-based all-sky-view camera observation},
JOURNAL = {Atmos. Meas. Tech.},
VOLUME = {13},
YEAR = {2020},
NUMBER = {4},
PAGES = {1953--1961},
DOI = {10.5194/amt-13-1953-2020}
}

@inproceedings{SWINySEG,
author = {Soumyabrata Dev and Florian M. Savoy and Yee Hui Lee and Stefan Winkler},
title = {{WAHRSIS: A low-cost high-resolution whole sky imager with near-infrared capabilities}},
volume = {9071},
booktitle = {Infrared Imaging Systems: Design, Analysis, Modeling, and Testing XXV},
editor = {Gerald C. Holst and Keith A. Krapels and Gary H. Ballard and James A. Buford Jr. and R. Lee Murrer Jr.},
organization = {International Society for Optics and Photonics},
publisher = {SPIE},
pages = {90711L},
year = {2014},
doi = {10.1117/12.2052982}
}

@ARTICLE{MPCM-Net,
  author={Niu, Penghui and She, Jiashuai and Cai, Taotao and Zhang, Yajuan and Zhang, Ping and Gu, Junhua and Li, Jianxin},
  journal={{IEEE} Trans. Geosci. Remote. Sens.}, 
  title={MPCM-Net: A multi-scale network that integrates partial attention convolution with Mamba for ground-based cloud image segmentation}, 
  year={2026},
  volume={},
  number={},
  pages={1-1},
  doi={10.1109/TGRS.2026.3666092}
}

@article{USF-Net,
  title={USF-Net: A Unified Spatiotemporal Fusion Network for Ground-Based Remote Sensing Cloud Image Sequence Extrapolation},
  author={Niu, Penghui and Cai, Taotao and She, Jiashuai and Zhang, Yajuan and Gua, Junhua and Zhanga, Ping and Hane, Jungong and Li, Jianxin},
  journal={arXiv preprint arXiv:2511.09045},
  year={2025},
  doi={https://doi.org/10.48550/arXiv.2511.09045}

}

@article{SENet,
  author       = {Jie Hu and
                  Li Shen and
                  Samuel Albanie and
                  Gang Sun and
                  Enhua Wu},
  title        = {Squeeze-and-Excitation Networks},
  journal      = {{IEEE} Trans. Pattern Anal. Mach. Intell.},
  volume       = {42},
  number       = {8},
  pages        = {2011--2023},
  year         = {2020},
  doi          = {10.1109/TPAMI.2019.2913372},
}

@inproceedings{ViT,
	author		= {Alexey Dosovitskiy and
                  Lucas Beyer and
                  Alexander Kolesnikov and
                  Dirk Weissenborn and
                  Xiaohua Zhai and
                  Thomas Unterthiner and
                  Mostafa Dehghani and
                  Matthias Minderer and
                  Georg Heigold and
                  Sylvain Gelly and
                  Jakob Uszkoreit and
                  Neil Houlsby},
	title		= {An Image is Worth 16x16 Words: Transformers for Image Recognition at Scale},
	booktitle	= {Proc. Int. Conf. Learn. Represent. (ICLR)},
	address		= {Virtual Event, Austria},
	month		= {May.},
	pages		= {3--7},
	year		= {2021},
	url          = {https://openreview.net/forum?id=YicbFdNTTy},
}

@inproceedings{Retractable,
  author       = {Jiale Zhang and
                  Yulun Zhang and
                  Jinjin Gu and
                  Yongbing Zhang and
                  Linghe Kong and
                  Xin Yuan},
  title        = {Accurate Image Restoration with Attention Retractable Transformer},
  booktitle    = {Int. Conf. Learn. Represent. (ICLR)},
  address	   = {Kigali, Rwanda},
  month		   = {May.},
  year         = {2023},
  url          = {https://openreview.net/forum?id=IloMJ5rqfnt}
}

@inproceedings{SViT,
  author       = {Ze Liu and
                  Yutong Lin and
                  Yue Cao and
                  Han Hu and
                  Yixuan Wei and
                  Zheng Zhang and
                  Stephen Lin and
                  Baining Guo},
  title        = {Swin Transformer: Hierarchical Vision Transformer using Shifted Windows},
  booktitle    = {Proc. IEEE. Int. Conf. Comput. Vision. (ICCV)},
  address	   = {Montreal, QC, Canada},
  month		   = {Oct.},
  pages        = {9992--10002},
  doi          = {10.1109/ICCV48922.2021.00986},
  year         = {2021}
}

@inproceedings{TimeMixer,
  author       = {Shiyu Wang and
                  Jiawei Li and
                  Xiaoming Shi and
                  Zhou Ye and
                  Baichuan Mo and
                  Wenze Lin and
                  Shengtong Ju and
                  Zhixuan Chu and
                  Ming Jin},
  title        = {TimeMixer++: {A} General Time Series Pattern Machine for Universal
                  Predictive Analysis},
  booktitle    = {Int. Conf. Learn. Represent. (ICLR)},
  address	   = {Singapore},
  month		   = {Apr.},
  year         = {2025},
  url          = {https://openreview.net/forum?id=1CLzLXSFNn}
}

@inproceedings{Vmamba,
  title={Vmamba: Visual state space model},
  author={Liu, Yue and Tian, Yunjie and Zhao, Yuzhong and Yu, Hongtian and Xie, Lingxi and Wang, Yaowei and Ye, Qixiang and Jiao, Jianbin and Liu, Yunfan},
  booktitle  = {Proc. Adv. neural inf. proces. syst. (NIPS)}, 
  address = {Vancouver, BC, Canada},
  month = {Dec.},
  pages    = {103031--103063},
  year         = {2024},
  url          = {http://papers.nips.cc/paper\_files/paper/2024/hash/baa2da9ae4bfed26520bb61d259a3653-Abstract-Conference.html},
}

@inproceedings{STT,
  author       = {Xiantao Hu and
                  Ying Tai and
                  Xu Zhao and
                  Chen Zhao and
                  Zhenyu Zhang and
                  Jun Li and
                  Bineng Zhong and
                  Jian Yang},
  title        = {Exploiting Multimodal Spatial-temporal Patterns for Video Object Tracking},
  booktitle    = {Proc. AAAI Conf. Artif. Intell. (AAAI)},
  address	   = {Philadelphia, PA, USA},
  month		   = {Feb.},
  pages        = {3581--3589},
  year         = {2025},
  doi          = {10.1609/AAAI.V39I4.32372}
}

@inproceedings{Sigma,
  author       = {Zifu Wan and
                  Pingping Zhang and
                  Yuhao Wang and
                  Silong Yong and
                  Simon Stepputtis and
                  Katia P. Sycara and
                  Yaqi Xie},
  title        = {Sigma: Siamese Mamba Network for Multi-Modal Semantic Segmentation},
  booktitle    = {Proc. IEEE Winter Conf. Appl. Comput. Vis. (WACV)},
  address	   = {Tucson, AZ, USA},
  month		   = {Mar.},
  pages        = {1734--1744},
  year         = {2025},
  doi          = {10.1109/WACV61041.2025.00176}
}

@inproceedings{U-Net,
	author    = {Olaf Ronneberger and
	Philipp Fischer and
	Thomas Brox},
	title     = {U-Net: Convolutional Networks for Biomedical Image Segmentation},
	booktitle = {Proc. Int. Conf. Med. Image Comput. Comput.-Assist. Intervent. (MICCAI)},
	address   = {Munich, Germany},
	month	= {Oct.},
	year      = {2015},
	pages     = {234--241},
	doi          = {10.1007/978-3-319-24574-4\_28}
}

@article{CloudFU-Net,
  author       = {Chaojun Shi and
                  Zibo Su and
                  Ke Zhang and
                  Xiongbin Xie and
                  Xian Zheng and
                  Qiaochu Lu and
                  Jiyuan Yang},
  title        = {CloudFU-Net: {A} Fine-Grained Segmentation Method for Ground-Based
                  Cloud Images Based on an Improved Encoder-Decoder Structure},
  journal      = {{IEEE} Trans. Geosci. Remote. Sens.},
  volume       = {62},
  pages        = {1--13},
  year         = {2024},
  doi          = {10.1109/TGRS.2024.3389089}
}

@inproceedings{SegFormer,
  author       = {Enze Xie and
                  Wenhai Wang and
                  Zhiding Yu and
                  Anima Anandkumar and
                  Jos{\'{e}} M. {\'{A}}lvarez and
                  Ping Luo},
  title        = {SegFormer: Simple and Efficient Design for Semantic Segmentation with
                  Transformers},
  booktitle    = {Proc. Adv. neural inf. proces. syst. (NIPS)}, 
  address = {Virtual, Online},
  month = {Dec.},
  pages        = {12077--12090},
  year         = {2021},
  url          = {https://proceedings.neurips.cc/paper/2021/hash/64f1f27bf1b4ec22924fd0acb550c235-Abstract.html}
}

@article{CloudSwinNet,
title = {CloudSwinNet: A hybrid CNN-transformer framework for ground-based cloud images fine-grained segmentation},
journal = {Energy},
volume = {309},
pages = {133128},
year = {2024},
issn = {0360-5442},
doi = {https://doi.org/10.1016/j.energy.2024.133128},
author = {Chaojun Shi and Zibo Su and Ke Zhang and Xiongbin Xie and Xiaoyun Zhang}
}

@article{LSTM,
  author={Hochreiter, Sepp and Schmidhuber, Jürgen},
  journal={Neural Comput.}, 
  title={Long Short-Term Memory}, 
  year={1997},
  volume={9},
  number={8},
  pages={1735-1780},
  doi={10.1162/neco.1997.9.8.1735}
}

@inproceedings{DLinear,
  author       = {Ailing Zeng and
                  Muxi Chen and
                  Lei Zhang and
                  Qiang Xu},
  title        = {Are Transformers Effective for Time Series Forecasting?},

  booktitle    = {Proc. AAAI Conf. Artif. Intell. (AAAI)}, 
  address = {Washington, DC, USA},
  month = {Feb.},
  pages        = {11121--11128},
  year         = {2023},
  doi          = {10.1609/AAAI.V37I9.26317}
}

@article{Timesnet,
  title={Timesnet: Temporal 2d-variation modeling for general time series analysis},
  author={Wu, Haixu and Hu, Tengge and Liu, Yong and Zhou, Hang and Wang, Jianmin and Long, Mingsheng},
  journal={arXiv preprint arXiv:2210.02186},
  year={2022},
  doi={https://doi.org/10.48550/arXiv.2210.02186}
}

@article{CNN-based,
  title={Convolutional neural networks for intra-hour solar forecasting based on sky image sequences},
  author={Feng, Cong and Zhang, Jie and Zhang, Wenqi and Hodge, Bri-Mathias},
  journal={Appl. Energy},
  volume={310},
  pages={118438},
  year={2022},
  doi={https://doi.org/10.1016/j.apenergy.2021.118438}
}

@article{ViT-based,
title = {A Transformer-based multimodal-learning framework using sky images for ultra-short-term solar irradiance forecasting},
journal = {Appl. Energy},
volume = {342},
pages = {121160},
year = {2023},
issn = {0306-2619},
doi = {https://doi.org/10.1016/j.apenergy.2023.121160},
author = {Jingxuan Liu and Haixiang Zang and Lilin Cheng and Tao Ding and Zhinong Wei and Guoqiang Sun}}

% % Here goes the biography details.

% % Here goes the biography details.

\end{document}